\documentclass{article}

\usepackage[nonatbib,preprint]{neurips_data_2021}

\usepackage[utf8]{inputenc} 
\usepackage[T1]{fontenc}    

\usepackage{times}
\usepackage{epsfig}
\usepackage{graphicx}
\usepackage{amsmath}
\usepackage{amsfonts}
\usepackage{amssymb}
\usepackage{microtype}
\usepackage{appendix}

\usepackage{epigraph}
\usepackage{paralist}
\usepackage{comment}
\usepackage{nicefrac}
\usepackage{tabularx}
\usepackage{booktabs}
\usepackage{array}
\usepackage{float}
\usepackage{multirow}
\usepackage{subcaption}
\usepackage{xspace}
\usepackage{siunitx}
\usepackage{bm} 
\usepackage[table,xcdraw]{xcolor}
\usepackage{capt-of}
\usepackage{caption}
\usepackage{relsize}
\usepackage[numbers,sort,compress]{natbib}
\usepackage{dirtytalk}
\usepackage{csquotes}
\usepackage[export]{adjustbox} 
\usepackage{placeins}

\usepackage[pagebackref=true,breaklinks=true,colorlinks,bookmarks=false]{hyperref}
\usepackage{url}
\usepackage{cleveref}
\usepackage{xspace}

\makeatletter
\DeclareRobustCommand\onedot{\futurelet\@let@token\@onedot}
\def\@onedot{\ifx\@let@token.\else.\null\fi\xspace}
\def\wrt{w.r.t\onedot}

\DeclareMathSymbol{@}{\mathord}{letters}{"3B}

\newcommand{\xhdr}[1]{\vspace{2pt}\noindent\textbf{#1}}
\newcommand{\code}[1]{\texttt{\small #1}}
\mathchardef\mhp="2D
\newcommand{\mhyphen}{~\mhp~}

\newcommand{\ms}[2]{$#1~\mathsmaller{\pm}~{\scriptstyle #2}$}
\newcommand{\bms}[2]{$\bm{#1}~\mathsmaller{\pm}~{\scriptstyle #2}$}
\newcommand{\res}[2]{{(\small #1,~#2)}}
\newcommand{\texthead}[1]{\noindent \textbf{#1}}

\newcolumntype{Y}{>{\centering\arraybackslash}X}
\newcolumntype{Z}{>{\centering\arraybackslash\hsize=.5\hsize}X}

\title{Habitat-Matterport 3D Dataset (HM3D): \\ 1000 Large-scale 3D Environments for Embodied AI}

\author{%
Santhosh K. Ramakrishnan$^{1,2}$,
Aaron Gokaslan$^{1, 5}$,
Erik Wijmans$^{1,3}$,
Oleksandr Maksymets$^{1}$,\\
\textbf{Alex Clegg$^{1}$,
John Turner$^{1}$,
Eric Undersander$^{1}$,
Wojciech Galuba$^{1}$,
Andrew Westbury$^{1}$,}\\
\textbf{Angel X. Chang$^{4}$,
Manolis Savva$^{4}$,
Yili Zhao$^{1}$,
Dhruv Batra$^{1,3}$}\\
$^{1}$Facebook AI Research~~$^{2}$UT Austin~~$^{3}$Georgia Tech~~$^{4}$Simon Fraser University~~$^{5}$Cornell University
}

\begin{document}

\maketitle

\begin{figure}[h]
\centering
\includegraphics[width=\linewidth,trim={0 5.0cm 0 0},clip]{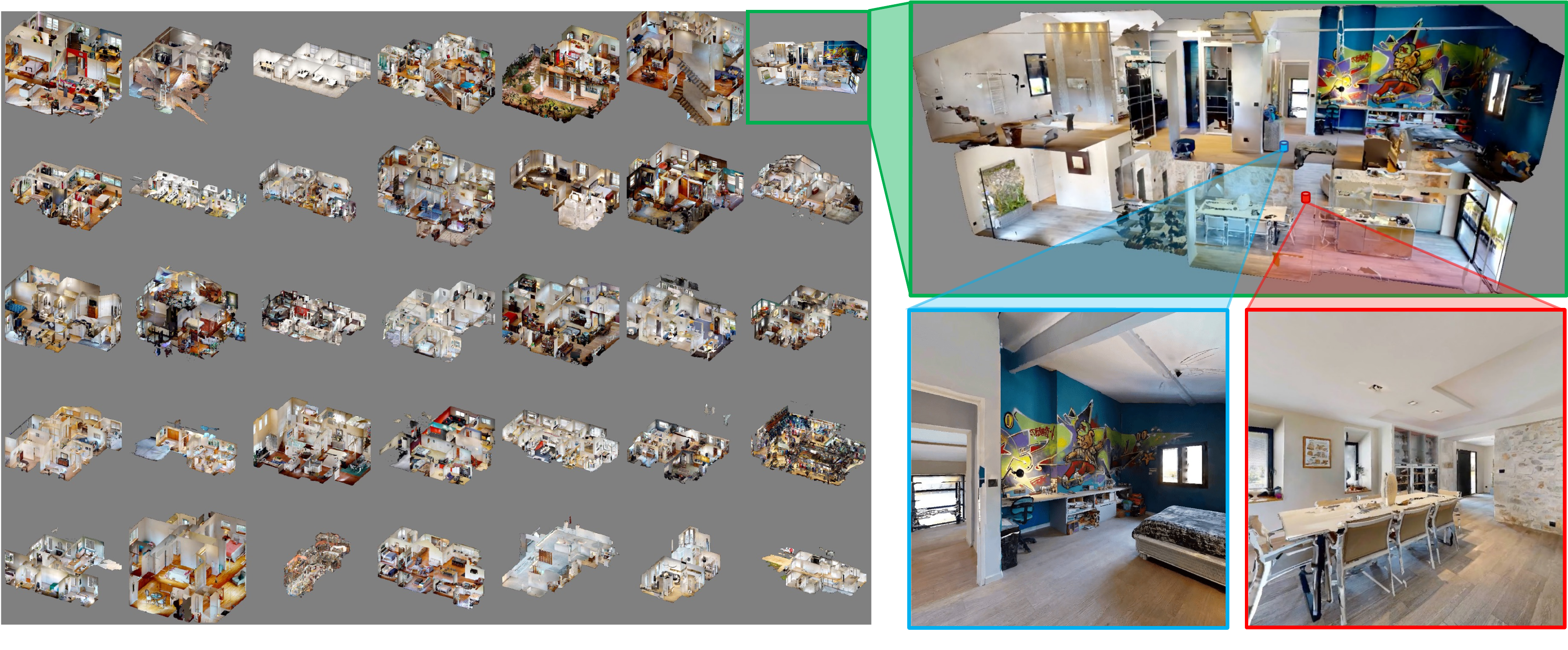}
\vspace*{-0.75cm}
\caption{\small The Habitat-Matterport 3D (HM3D) dataset of large-scale 3D and photorealistic environments provides $1@000$ building-scale reconstructions of interiors from a diverse set of geographic locations. The scale, completeness, and visual fidelity of these reconstructions surpass those of prior datasets, and enable research on embodied AI agents that can perceive, navigate, and act within realistic indoor environments. The image on the left displays a collage of a subset of HM3D scans. The image on the top-right is a close-up view of a specific scan, and the images on the bottom-right are snapshots from two camera viewpoints in the scan.}
\label{fig:teaser}
\end{figure}

\begin{abstract}
We present the Habitat-Matterport 3D (HM3D) dataset.
HM3D is a large-scale dataset of $1@000$ building-scale 3D reconstructions from a diverse set of real-world locations.
Each scene in the dataset consists of a textured 3D mesh reconstruction of interiors such as multi-floor residences, stores, and other private indoor spaces. \hfill\vspace{0.15cm} \\
HM3D surpasses existing datasets available for academic research in terms of physical scale, completeness of the reconstruction, and visual fidelity. HM3D contains $112.5k~\si{m^2}$ of navigable space, which is $1.4 \mhyphen 3.7 \times$ larger than other building-scale datasets such as MP3D and Gibson. When compared to existing photorealistic 3D datasets such as Replica, MP3D, Gibson, and ScanNet, images rendered from HM3D have $20 \mhyphen 85\%$ higher visual fidelity \wrt counterpart images captured with real cameras, and
HM3D meshes have $34 \mhyphen 91\%$ fewer artifacts due to incomplete surface reconstruction. \hfill\vspace{0.15cm} \\
The increased scale, fidelity, and diversity of HM3D directly impacts the performance of embodied AI agents trained using it. In fact, we find that HM3D is `pareto optimal' in the following sense -- agents trained to perform PointGoal navigation on HM3D achieve the highest performance regardless of whether they are evaluated on HM3D, Gibson, or MP3D. No similar claim can be made about training on other datasets. HM3D-trained PointNav agents achieve $100\%$ performance on Gibson-test dataset, suggesting that it might be time to retire that episode dataset. 
\end{abstract}

\section{Introduction}

As we seek to develop intelligent AI agents that can assist us in our daily activities, good models of indoor 3D environments are becoming increasingly important.
Consequently, recent years have seen growing demand for datasets of 3D interiors, whether acquired from the real world, or authored by artists using 3D design tools.
Scene datasets based on real-world interiors can be used to develop and evaluate computer vision systems (e.g., on object detection and semantic segmentation tasks), or to train AI agents to navigate and follow instructions in an embodied setting.
The latter research agenda in particular has been accelerated by the availability of realistic 3D datasets and high-performance 
simulators that dramatically reduce the time and logistical complexity for developing AI agents.

Unfortunately, there are only a handful of datasets of indoor 3D environments captured from the real world.
Early efforts on 3D scene datasets such as SceneNN~\cite{hua2016scenenn} and ScanNet~\cite{dai2017scannet} collected reconstructions of regions of rooms, and individual rooms.
Other datasets that provide 3D reconstructions of entire buildings such as the BuildingParser~\cite{armeni2017joint}, Matterport3D~\cite{chang2017matterport3d} and Gibson~\cite{xiazamirhe2018gibsonenv} efforts are either limited in total size or suffer from incomplete reconstructions.

We present the Habitat-Matterport 3D Dataset (HM3D), a large dataset of building-scale reconstructions of a diverse set of real-world spaces.
HM3D provides $1@000$ near-complete high-fidelity reconstructions of entire buildings (see \Cref{fig:teaser}).
Each of these reconstructions provides a capture of the habitable and navigable space of each interior.
In total, the dataset contains more than $10@600$ rooms across approximately $1@920$ building floors with a navigable area of $112.5k~\si{m^2}$.
The real-world interiors from which these reconstructions are acquired span a diverse set of categories (eg. multi-floor residences, offices, restaurants, and shops), geographical locations, and physical sizes.

Three key characteristics distinguish HM3D relative to prior work on real-world scanned indoor 3D datasets: \emph{scale}, \emph{completeness}, and \emph{visual fidelity}.
Unlike prior datasets, each scene in HM3D typically represents a complete building such as a multi-floor private residence. 
Therefore, HM3D has significantly higher total navigable area ($1.4 \mhyphen 3.7 \times$ larger), which is particularly important for embodied AI tasks such as navigation.
The completeness of HM3D is reflected in $34 \mhyphen 91\%$ reduction in reconstruction artifacts due to missing surfaces, holes, or untextured surface regions when compared to prior photorealistic 3D datasets. 
This increased surface completeness leads to lower incidence of highly unrealistic `seeing through a hole in the wall' issues that can be detrimental to embodied AI agent training.
Finally, the visual fidelity of images rendered from HM3D is $20 \mhyphen 85\%$ higher than prior large-scale datasets, which can help to train better embodied AI agents that generalize to real-world settings.
As the name suggests, HM3D is `Habitat-ready', meaning that it comes prepacked with meta-data and support necessary to be used with the Habitat simulator~\cite{savva2019habitat} for training embodied AI agents to understand and navigate 3D spaces.

We carry out a number quantitative analyses and experiments to understand the characteristics of HM3D.
First, we compare rendered images from HM3D and other 3D scan datasets to camera-captured images from the counterpart real-world interiors, and find that HM3D has significantly higher visual fidelity than other datasets.
Second, we find that HM3D has fewer artifacts leading to incompleteness and `holes' in surface reconstruction.
Finally, we train agents for the task of PointGoal navigation~\cite{anderson2018evaluation} using HM3D and other datasets, and find that agents trained in HM3D generalize well across environments. In particular, HM3D is pareto-optimal in the sense that HM3D-trained agents achieve the best performance across Gibson, MP3D, and HM3D test sets. HM3D-trained agents also achieve perfect success on Gibson test scenes, and obtain $3 \mhyphen 4$ points higher success and SPL on MP3D test scenes when compared to the next-best agent. These results strongly suggest that embodied agents benefit from the increased scale and diversity of HM3D.
\section{Related Work}

3D datasets can broadly be categorized into synthetic/CAD-based, 3D reconstruction or mesh-based, floorplan-based and panorama-based datasets.

\xhdr{Synthetic 3D scene datasets.}
Embodied AI simulation engines often make use of synthetic scenes with rearrangeable objects~\cite{kolve2017ai2,yan2018chalet,puig2018virtualhome,szot2021habitat}.
Often these scenes are limited to isolated rooms or individual rooms that are connected via a magic portal~\cite{yan2018chalet}.
There are also datasets of building-scale synthetic scenes~\cite{song2017semantic,fu20203dfront,szot2021habitat}.
However, these authored scenes often do not reflect the variety of architectural layout as well as object arrangement and clutter in the real world. 
Typically, objects in datasets of synthetic scenes are limited in visual and geometric diversity, since the same set of objects are reused across scenes.
In addition, there is a sim-to-real gap between the rendered appearance of synthetic objects and real-world objects.
To limit this discrepancy between synthetic and real domains, there are a number of recent synthetic scene dataset efforts that are designed from real world counterpart environments~\cite{deitke2020robothor,szot2021habitat}.
HM3D is a reconstruction dataset capturing the layout and appearance of a large number of real buildings.

\xhdr{3D reconstruction datasets.}
Existing reconstructions of indoor spaces are limited in scale.
Common reconstruction datasets consist primarily of scans for regions of rooms and single rooms~\cite{hua2016scenenn,dai2017scannet,straub2019replica,halber2019rescan,wald2019rio}.\footnote{ScanNet and Replica do contain a number of multi-room scenes.}
There exist datasets with building level reconstruction, but these are limited in the overall number of scenes and real-world spaces (BuildingParser~\cite{armeni20163d}, 2D-3D-S~\cite{armeni2017joint}, Matterport3D~\cite{chang2017matterport3d}).
The largest building-level reconstruction dataset is Gibson~\cite{shen2020igibson} which consists of $571$ scenes. However, many scans from Gibson suffer from reconstruction artifacts and `holes' due to partially reconstructed surfaces. Prior work performed manual inspection and found that only $106$ / $571$ Gibson scans are of acceptable quality (i.e., $\ge 4$ on a scale of 0-5)~\cite{savva2019habitat}.
\citet{dehghan2021arkitscenes} have recently released reconstructions of $1@661$ scenes but most are single room-scale regions, from rented homes in three European cities.
HM3D contains $1@000$ building-scale reconstructions spanning a diverse set of locations around the world.

\xhdr{Floorplan and panorama datasets.}
Floorplan datasets~\cite{lifullhome,kalervo2019cubicasa5k,wu2019data} can be converted to 3D floorplans outlining the architectural layout of buildings and rooms using heuristics.
However, the architectural layout tends to be oversimplified as there is typically no specification of wall height or ground level (i.e. all rooms have equal height and simple flat ceilings).
Most importantly, these datasets do not provide the textured appearance of the environments or of furniture and other objects present in the rooms.
Recently, the Zillow indoor dataset~\cite{cruz2021zillow} provides floorplans and panorama images captured from a variety of properties.
However, almost all captured properties are unfurnished, and even with panorama images available it is not easy to produce 3D mesh reconstructions of the interiors.
In contrast, HM3D provides a large number of interiors that are reconstructed to a higher surface completeness and with higher visual fidelity than prior reconstruction datasets.

\begin{figure}[t]
\includegraphics[width=\linewidth, trim={0 8.5cm 6.5cm 0},clip]{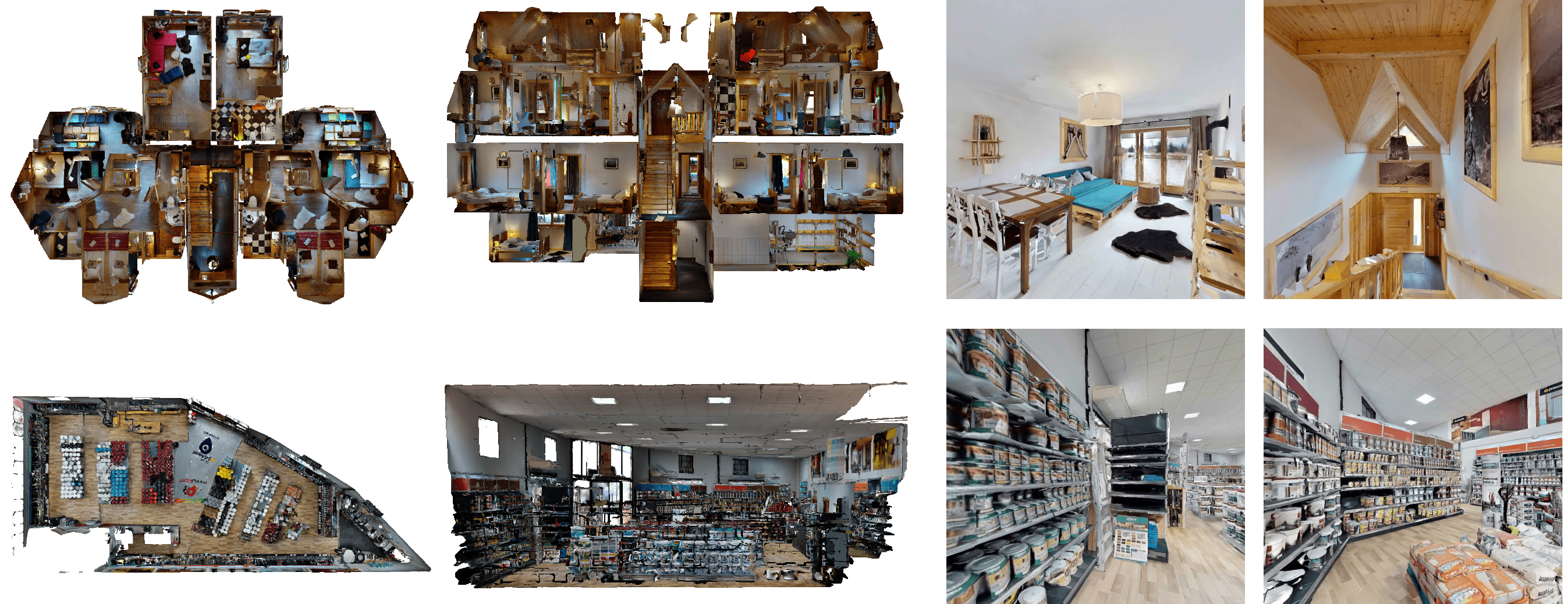}

\caption{\small Two example scenes from the HM3D dataset. From left to right in each row: top-down view, cross section view, and two egocentric views from navigable positions in the scene. The dataset contains a wide range of environments such as residences, stores, and workplaces. See Appendix~\ref{suppsec:examples} for more examples.}
\label{fig:scene-examples}
\end{figure}

\begin{figure}
    \centering
    \includegraphics[width=0.30\linewidth]{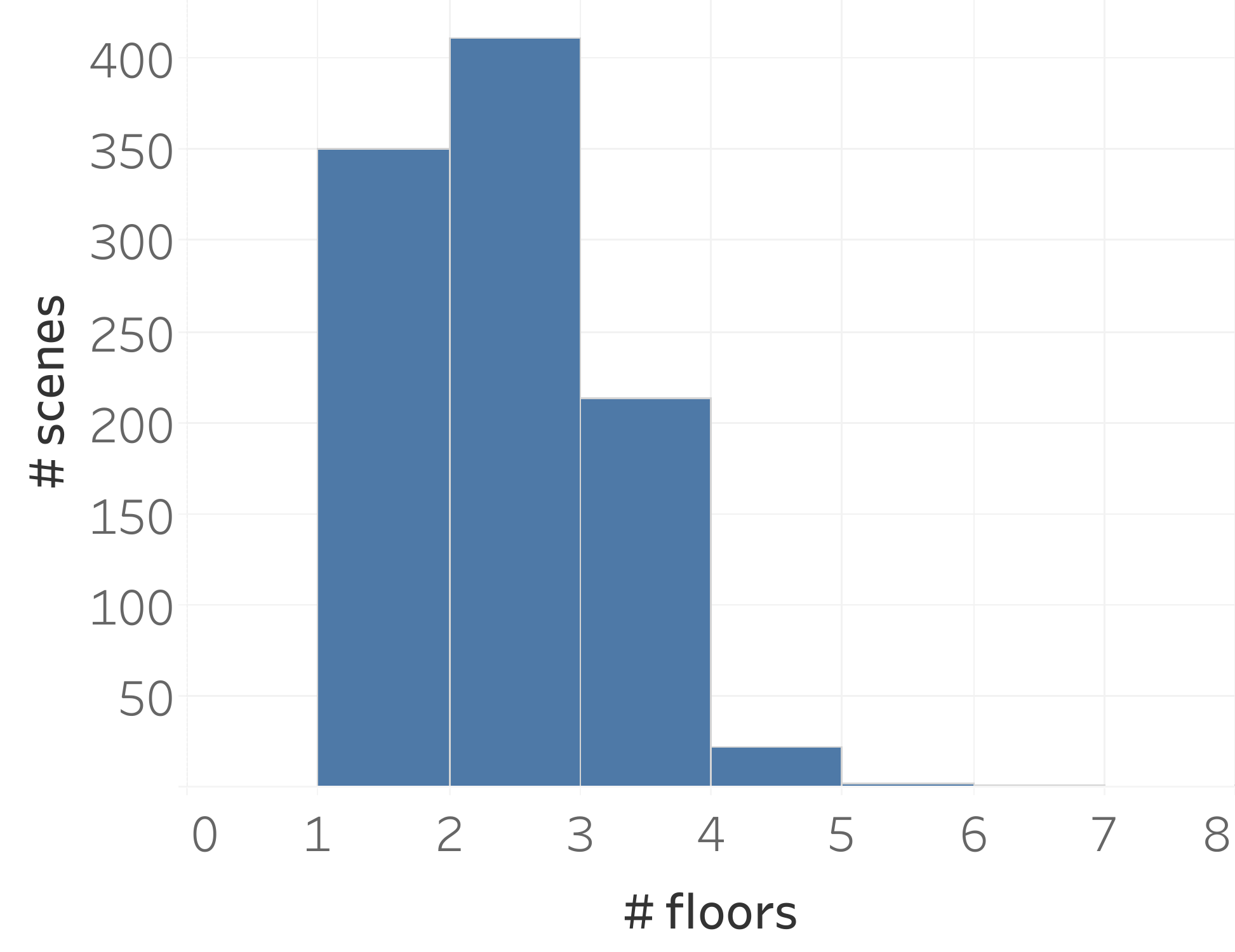}
    \includegraphics[width=0.30\linewidth]{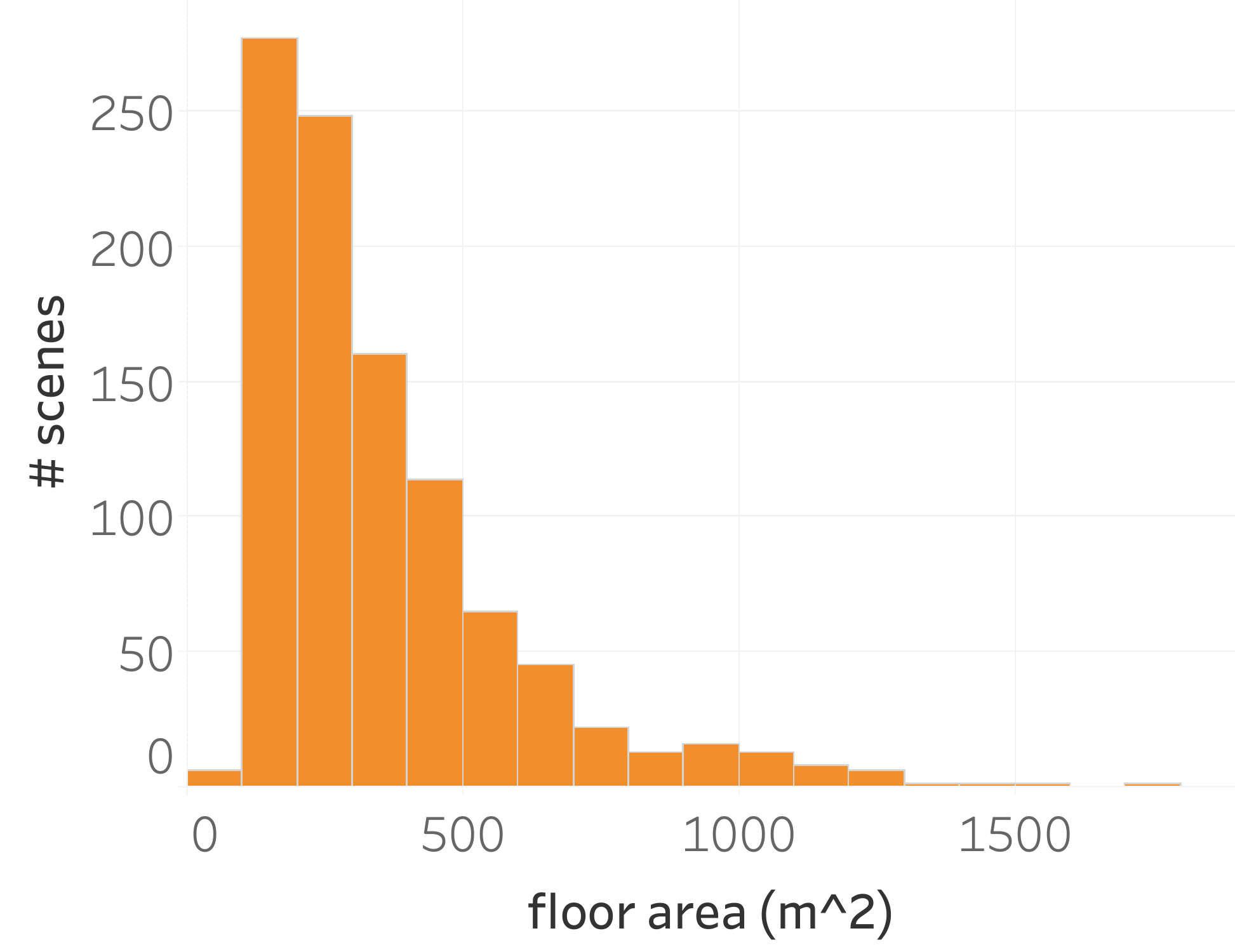}
    \includegraphics[width=0.30\linewidth]{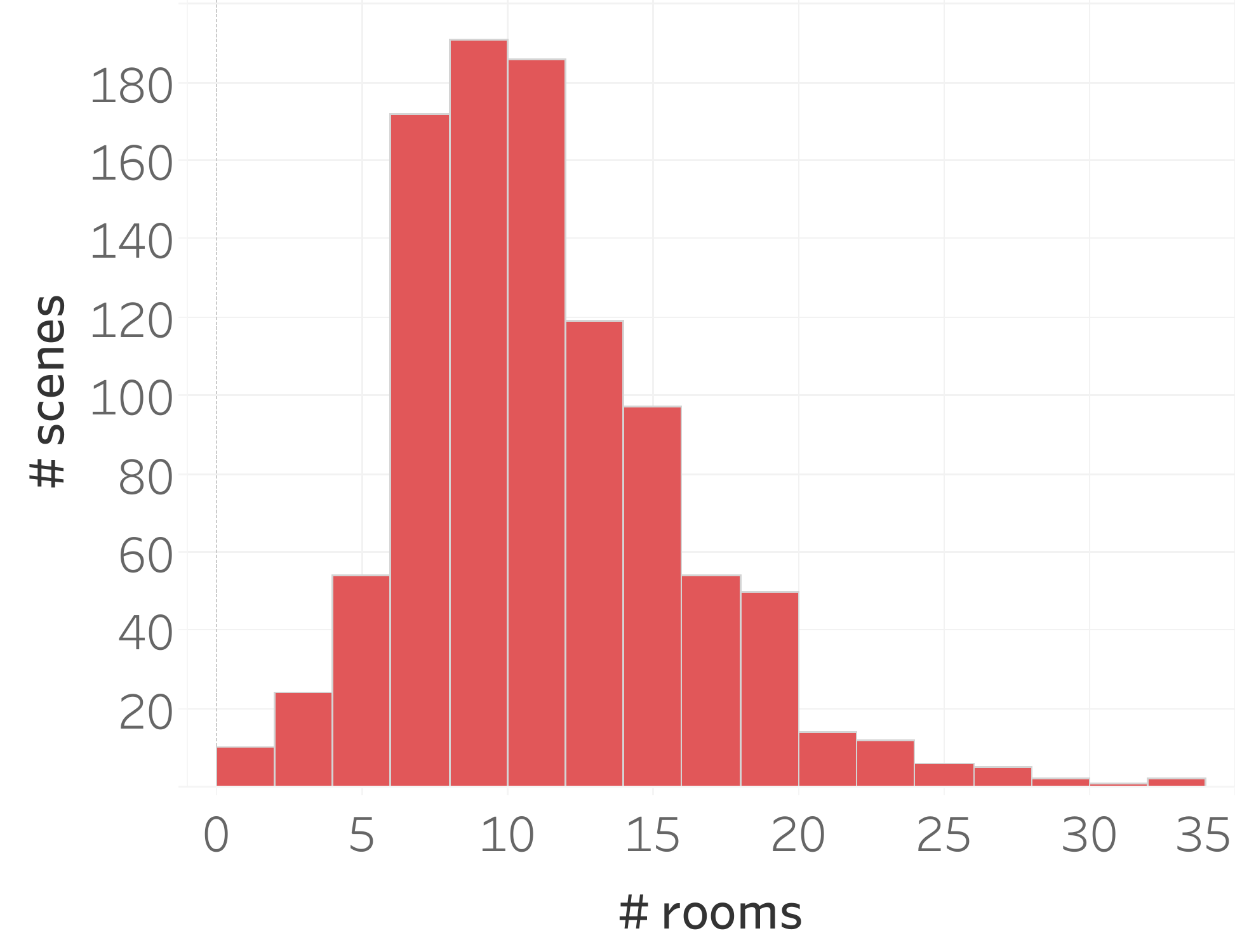}
    \caption{\small From left to right: i) histogram of distribution over number of floors per scene; ii) histogram of total floor area; iii) histogram of distribution over number of rooms per scene. HM3D scenes span a broad spectrum of physical scale.}
    \label{fig:floor-room-distributions}

\end{figure}
\section{Dataset}

The Habitat-Matterport 3D Dataset (HM3D) is a collection of $1@000$ 3D reconstructions and consists of multi-floor residences, stores, and other private indoor spaces.
All spaces were scanned using a Matterport Pro2 tripod-based depth sensor.\footnote{\url{https://matterport.com/cameras/pro2-3D-camera}}
Alignment of the RGB-D data, surface meshing, and texturing were carried out using the reconstruction pipeline provided by Matterport, Inc.
Scans were taken from spaces in $38$ countries, and $181$ geographic regions (states, provinces etc.) across those countries.
In the United States, spaces are located in $43$ states.
\Cref{fig:scene-examples} shows some example scenes.

The set of $1@000$ scenes in HM3D were curated from a larger pool of candidate scenes through a two-stage annotation and verification process.
First, a group of 15 volunteer annotators rated each scene on a 1-5 quality scale assessing scene quality.
The annotators visually inspected each scene for reconstruction artifacts such as holes/cracks, the presence of realistic and dense furnishing, the number of closed doors (which prevent access to some rooms), and the `interactive potential' of the scene (based on objects with which a person might interact).
The annotators also had access to quantitative metrics characterizing the navigability of the scene so that they could detect reconstruction issues causing disconnectedness in the building floors.
After a round of rating, a second set of three expert annotators collated scenes by first sorting highly ranked scenes and then selecting so as to preserve diversity in the number of floors per scene.
In total, HM3D curation, annotation, and release represents an estimated $800$+ hours of human effort.

The final set of scenes spans a broad spectrum of total area, with the smallest scene having a floor area of $49\text{m}^2$ and the largest scene an area of $2@172\text{m}^2$.
The architectural layout of the scenes also spans a broad spectrum with buildings of between one and eight floors, and between one `room' and $93$ rooms.\footnote{Room statistics were obtained using the mesh chunk meta-data from the Matterport reconstruction pipeline. Each mesh chunk is created by the reconstruction pipeline from a set of tripod locations in the same room.}
More detailed statistics regarding the dataset composition are visualized in \Cref{fig:floor-room-distributions}.

\subsection{Scale comparison}

We compare the scale of HM3D to other datasets using a number of metrics that measure the overall floor area, navigable area, and structural complexity of the scenes.

\xhdr{Floor area ($\si{m^2}$)} measures the overall extents of the floor regions in the scene.
This is the area of the 2D convex hull of all navigable locations in a floor.
For scenes with multiple floors, the floor space is summed over all floors.
This is implemented in the same way as by \citet{xiazamirhe2018gibsonenv} to make the reported statistics comparable.
Higher values indicate the presence of more navigation space and rooms.

\xhdr{Navigable area ($\si{m^2}$)} measures the total scene area that is actually navigable in the scene.
This is computed for a cylindrical robot with radius $0.1\si{m}$ and height $1.5\si{m}$ using the AI Habitat~\cite{savva2019habitat} navigation mesh implementation.
This area is strictly lower than the floor area as it excludes points that are not reachable by the robot.
Higher values indicate larger quantity and diversity of viewpoints for a robot. 

\xhdr{Navigation complexity} measures the difficulty of navigating in a scene.
This is computed as the maximum ratio of geodesic path to euclidean distances between any two navigable locations in the scene.
This is the same metric as reported for the original Gibson dataset to again make the statistics comparable~\cite{xiazamirhe2018gibsonenv}.
Higher values indicate more complex layouts with navigation paths that deviate significantly from straight-line paths.

\xhdr{Scene clutter} measures the amount of clutter in the scene.
This is computed as the ratio between the raw scene mesh area within $0.5\si{m}$ of the navigable regions and the navigable space.
Higher values are better and indicate more cluttered scenes that provide more obstacles for navigation.

\Cref{tab:dataset-comparison} reports the values of these metrics for HM3D as well as a number of other indoor datasets, primarily focusing on existing 3D reconstruction datasets.
We also compute the metrics for the RoboTHOR~\cite{deitke2020robothor} dataset which is synthetic but based on real-world layouts.
The chosen comparison points span a spectrum of total sizes and complexities.
For Gibson, note a second set of metric values for the restricted subset of fewer ``high quality'' Gibson scenes that were rated as at least 4/5 by a set of human annotators.
This subset of Gibson exhibits fewer reconstruction artifacts than the full Gibson dataset (see \citet{savva2019habitat} for a description of the original rating process).

We can make a number of observations.
First, HM3D provides $1.7\times$ higher floor area and $1.4\times$ higher navigable area compared to the Gibson (previously the largest). In particular, HM3D provides $20\times$ higher floor area and $15.6\times$ higher navigable area if we only consider the high quality reconstructions in Gibson 4+.
Second, the scene clutter of HM3D exceeds that of most other datasets by $\sim\!\!1.2\times$ with the exception of RoboTHOR which is a significantly smaller dataset.
Finally, the navigation complexity metric shows that HM3D scenes are relatively complex to navigate, close to other building-scale datasets such as MP3D and Gibson (by a factor of $0.8 \mhyphen 1.1\times$), and higher than room-scale datasets such as Replica, RoboTHOR and ScanNet (by a factor of $2.2 \mhyphen 6.4\times$).

\begin{table*}[t]
\centering
\resizebox{\textwidth}{!}{
\begin{tabular}{@{}lcccccc@{}}
\toprule
Dataset                    &   Replica~\cite{straub2019replica}      &   RoboTHOR~\cite{deitke2020robothor}     &     MP3D~\cite{chang2017matterport3d}        &    Gibson~\cite{xiazamirhe2018gibsonenv} (4+ only)   &    ScanNet~\cite{dai2017scannet}    &     HM3D (ours)     \\ \midrule
Number of scenes           &     $18$       &     $75$       &     $90$        &      $571~(106)$      &    $1613$     &     $1000$          \\
Floor area (\si{m^2})     &     $2.19k$    &     $3.17k$    &   $101.82k$     &  $217.99k~(17.74k)~~$ &   $39.98k$    &     $365.42k$       \\
Navigable area (\si{m^2}) &     $0.56k$    &     $0.75k$    &    $30.22k$     &     $81.84k~(7.18k)~~$&    $10.52k$   &     $112.50k$       \\
Navigation complexity      &     $5.99$     &     $2.06$     &    $17.09$      &     $14.25~(11.90)$   &    $3.78$     &      $13.31$        \\
Scene clutter              &      $3.4$     &      $8.2$     &     $2.99$      &      $3.14~(3.04)$    &    $3.15$     &      $3.90$         \\ \bottomrule
\end{tabular}
}
\caption{
Comparison of HM3D to other existing indoor scene datasets.
Gibson 4+ refers to the subset of Gibson scenes that were rated as ``high quality'' and relatively free of reconstruction errors~\cite{savva2019habitat}.
HM3D surpasses previously available datasets with $1.8\times$ more scans, $1.6 \mhyphen 3.6\times$ higher total physical size, provides $1.2\times$ more cluttered scenes, and has relatively high navigation complexity.\vspace{-0.2cm}
}
\label{tab:dataset-comparison}
\end{table*}

\subsection{Reconstruction completeness comparison}
\label{sec:rec_compl}
\begin{figure*}[t]
    \centering
    \includegraphics[width=\linewidth, trim={0 5.5cm 0.0cm 0},clip]{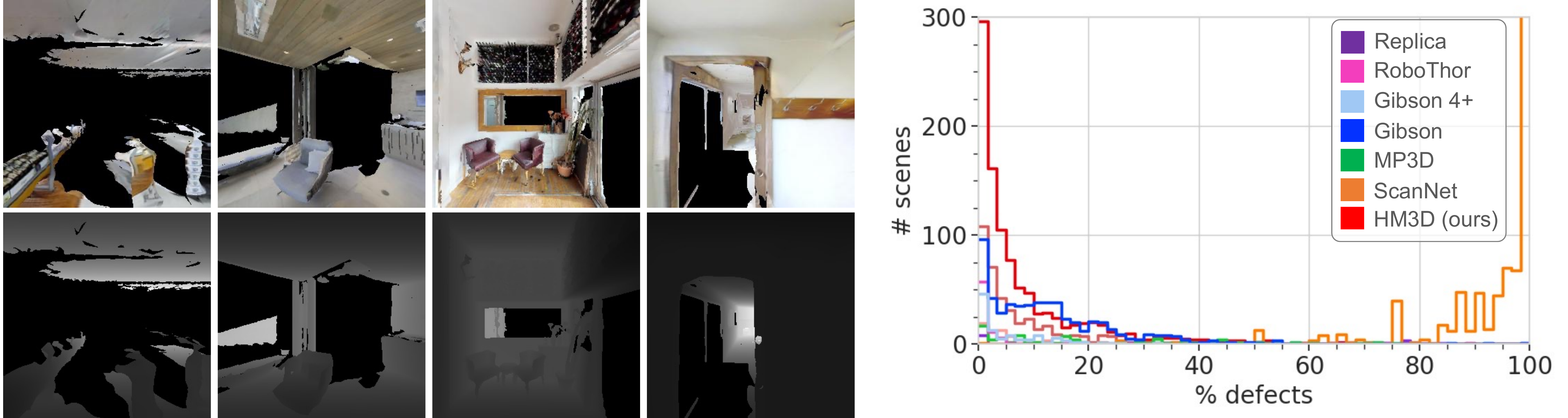}
    \caption{\small We measure the incidence of reconstruction artifacts such as missing surfaces, holes, or untextured surface regions (`defects') using a view-based metric. \textbf{Left:} Examples of viewpoints with significant mesh reconstruction artifacts. \textbf{Right:} Histogram of defect metric over the scenes from different datasets. The X-axis measures the fraction of densely sampled views in a scene that exhibit significant defects. HM3D provides the largest number of scans with minimal reconstruction artifacts.\vspace{-0.2cm}
    }
    \label{fig:example-defects}
\end{figure*}

When reconstructing a 3D mesh from scanned images, it is common to encounter reconstruction artifacts (or defects) such as missing surfaces, holes, or 
untextured surface regions. These reconstruction artifacts lower the visual quality of rendered images and may increase
the domain gap between learning within the simulator and deploying to the real world. 
HM3D offers more complete reconstructions with fewer instances of missing surfaces or reconstruction artifacts resulting in `holes' and `black cracks'.
We design a view-based metric to measure the degree to which such artifacts occur in HM3D and other datasets. 
First, we densely sample camera viewpoints in a scene as follows. We divide the set of navigable locations in a scene into a grid with $1\si{m} \times 1\si{m}$ cells.
At each grid location, we sample 4 camera viewpoints by varying the agent's heading angle ($0^\circ, 90^\circ, 180^\circ, 270^\circ$) and fixing the azimuth angle to $0^\circ$ (i.e., agent looks straight ahead).
Finally, we place a $90^\circ$ field-of-view RGB-D camera at each viewpoint and render $256 \times 256$ images.
This sampling process is uniform (in terms of coverage of the floor space) and adaptive (i.e., the number of images varies with the size of the scene).
For each viewpoint, we compute the fraction of depth pixels with depth value of $0$ (i.e., invalid depth pixels) and the fraction of RGB pixels that are completely black (i.e. RGB value of $0$).\footnote{The pixels corresponding to missing surfaces / textures are set to $0$ for both RGB and depth rendering.}
Any views with more than $5$\% of such pixels either in the depth frame or the RGB frame are marked as exhibiting an artifact.
\Cref{fig:example-defects} (left) shows examples of views with significant defects.
In reconstructions that are complete and that do not exhibit holes and cracks, there will be fewer such views.
We summarize this metric by computing the proportion of sampled views in a scene that exhibit such significant reconstruction artifacts to arrive at an overall `\% defects' value for the scene.

We compare the distribution of this metric for HM3D and other datasets in \Cref{fig:example-defects} (right).
Overall, we see that HM3D scenes exhibit fewer artifacts (more scenes with lower `\% defect' values).
While ScanNet offers more scans than HM3D, almost all scans from ScanNet exhibit severe reconstruction artifacts.
Other large-scale datasets such as Gibson, and MP3D exhibit broader distributions with a significant number of scenes having fairly high reconstruction defect values.
HM3D has more than three times as many scenes with less than 5\% of views exhibiting artifacts compared to Gibson (560 scenes vs 175 scenes).
As expected, Gibson 4+ provides a smaller but higher-quality subset of Gibson scenes with 
fewer reconstruction artifacts.
While RoboTHOR scans have very few artifacts, they are not photorealistic and are small in quantity.
The Replica scans are much smaller in number, and some scans have $\sim\!80\%$ defects since the roof is missing.
Overall, HM3D offers the largest number of scans with high completeness.

\subsection{Visual fidelity comparison}
\label{sec:visual_fidelity}

\begin{figure}[t]
    \centering
    \begin{subfigure}[b]{0.37\textwidth}
        \centering
        \includegraphics[width=\linewidth, trim={0 3cm 15.0cm 0},clip]{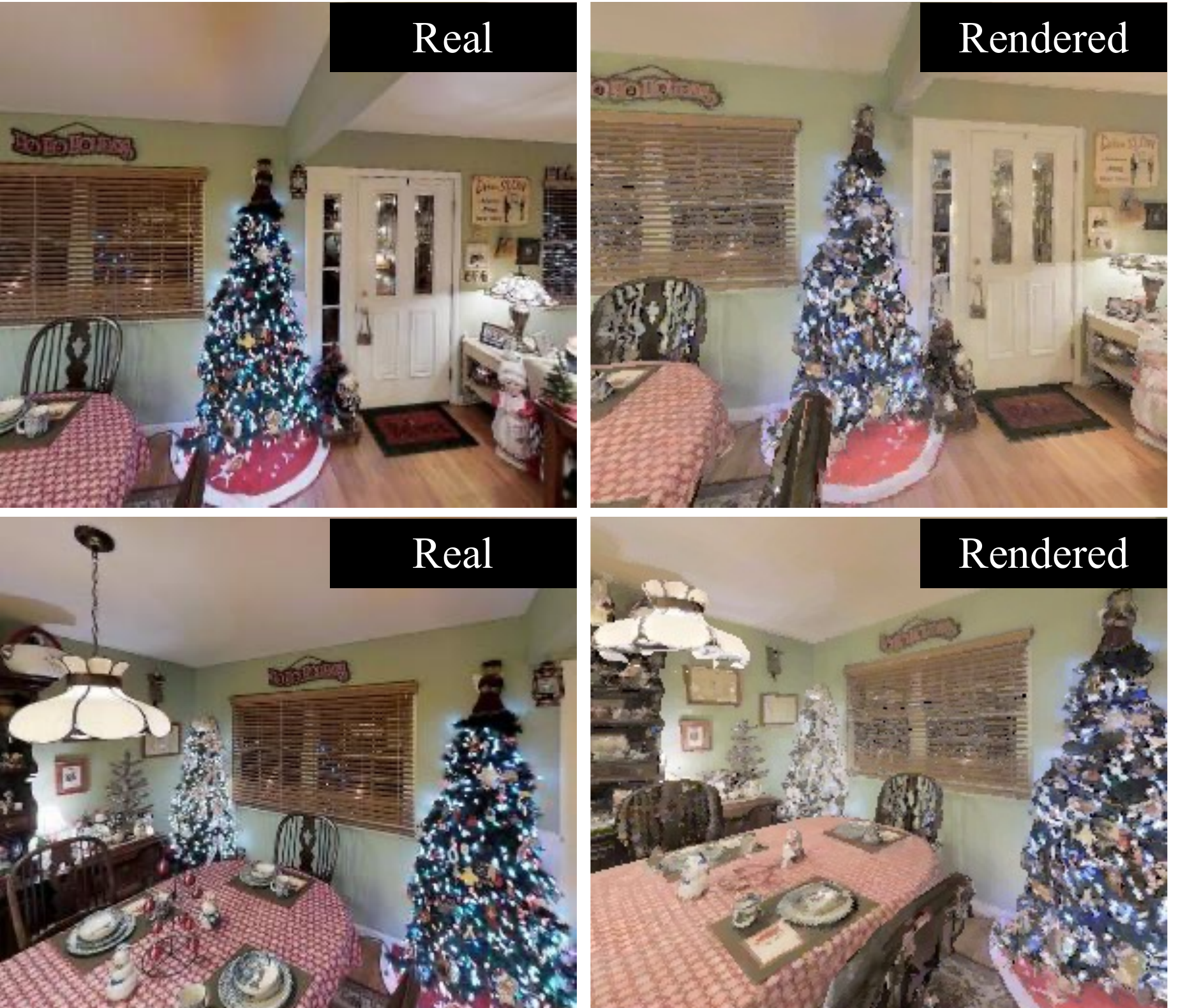}
        \caption{\small Real vs. rendered images.}
    \end{subfigure}
    \hspace{0.1cm}
    \begin{subtable}[b]{0.60\textwidth}
        \centering
        \resizebox{1.0\linewidth}{!}{%
        \begin{tabular}{@{}lrrrrr@{}}
        \toprule
                      &             & \multicolumn{2}{c}{Gibson real}                     & \multicolumn{2}{c}{MP3D real}                   \\
        \cmidrule(lr){3-4} \cmidrule(lr){5-6}
        Dataset       & $\#$ scenes & FID $\downarrow$  & KID{\small $\times 10^3$} $\downarrow$ & FID $\downarrow$ & KID{\small $\times 10^3$} $\downarrow$ \\ \midrule
        Replica       & $18$        & $34.9$            &\bms{15.8}{0.8}                         & $42.8$           & \ms{19.3}{0.8}                         \\
        RoboTHOR      & $75$        & $157.6$           & \ms{109.8}{1.8}                        & $163.0$          & \ms{111.3}{1.8}                        \\
        MP3D          & $90$        & $43.8$            & \ms{32.9}{1.6}                         & $24.4$           & \ms{17.3}{1.2}                         \\
        Gibson 4+     & $106$       & $27.4$            & \ms{18.9}{0.8}                         & $32.6$           & \ms{20.0}{0.8}                         \\
        Gibson        & $571$       & $39.3$            & \ms{29.8}{1.5}                         & $38.0$           & \ms{25.5}{1.0}                         \\
        ScanNet       & $1613$      & $126.7$           & \ms{106.0}{2.4}                        & $121.3$          & \ms{97.4}{2.30}                        \\
        HM3D          & $1000$      & $\bm{20.5}$       &\bms{15.8}{1.0}                         & $\bm{20.5}$      &\bms{12.8}{0.8}                         \\ \midrule
        MP3D real     & $90$        & $11.2$            & \ms{6.2}{0.7}                          & $0.0$            & \ms{0.0}{0.1}                          \\
        Gibson real   & $571$       & $0.0$             & \ms{0.0}{0.1}                          & $11.2$           & \ms{6.2}{0.7}                          \\ \bottomrule
        \end{tabular}
        }
        \caption{\small Visual fidelity comparison of rendered images with real images.\vspace{-0.1cm}}
    \end{subtable}
    \caption{\small Visual fidelity comparison of HM3D and other reconstruction datasets. We render images from the reconstructed scenes in each dataset using Habitat~\cite{savva2019habitat}, and extract real RGB images from raw panoramas in Gibson and MP3D (see comparison on the left). Then, we compute the FID and KID of the rendered images by comparing them against the real images. Lower values indicate closer distributional match to the statistics of the real image sets. HM3D provides significantly lower FID and KID than images rendered from other datasets, even in the case of images rendered from Gibson reconstructions evaluated against Gibson real images. This result indicates the high visual fidelity of HM3D reconstructions relative to other datasets.\vspace{-0.2cm}}
    \label{tab:visual_quality}
\end{figure}

We also compare the overall visual quality of rendered images from HM3D with prior datasets.
For each dataset, we use the RGB images from \Cref{sec:rec_compl} to ensure that we assess the visual fidelity of rendered images from all parts of a scene.
We compare the image quality against a set of real RGB images generated from high-resolution panoramas (i.e., $360^\circ$ field-of-view equirectangular images) in Gibson and MP3D using the FID~\cite{heusel2017gans} and KID~\cite{binkowski2018demystifying} metrics.
We refer to these sets of real RGB images as `Gibson real' and `MP3D real'.
\Cref{tab:visual_quality} summarizes the results of this comparison.

The quality of images rendered from HM3D is much closer to real images when compared to the other datasets.
Out of all datasets, images rendered from HM3D exhibit the lowest FID / KID scores when compared with both MP3D real ($20.53 / 15.78$)
and Gibson real ($20.49 / 12.76$).
Note that we observe a domain shift between datasets that leads to non-zero FID and KID scores even for real images. Comparing images from Gibson real with MP3D real provides a `lower bound' of $6.16 - 6.23$ against which we can compare the metric values for rendered dataset images.
As expected, images rendered from RoboTHOR have the highest FID and KID scores since they are not photorealistic. Images rendered from ScanNet also have high FID and KID scores since they exhibit significant mesh artifacts (see Sec.~\ref{sec:rec_compl}). Images rendered from Replica have relatively high distance scores (despite having high quality scans) due to the lack of textured ceilings in multiple scans (eg., $34.94$ FID vs. Gibson real, and $42.76$ FID / $19.31$ KID vs. MP3D real). Images rendered from the remaining datasets have significantly higher FID and KID values when compared to HM3D, showing that they have lower visual fidelity as measured against the real images from Gibson and MP3D.

\section{Experiments}

A popular downstream application for large-scale 3D reconstruction datasets has been to use them with 3D simulation platforms~\cite{savva2019habitat,shen2020igibson,deitke2020robothor} to study embodied AI tasks such as visual navigation~\cite{gupta2017cognitive,savinov2018semi,savva2019habitat,wijmans2019dd,chaplot2020neural}.
As described in the previous sections, HM3D improves over existing datasets both in terms of size and quality.
In this section, we perform experiments to show that navigation agents trained on HM3D benefit from its scale and quality, and generalize better when transferred to other datasets.

\begin{table*}[t]

\begin{minipage}{\linewidth}
    \centering
    \includegraphics[width=\textwidth,trim={0 5.0cm 4.5cm 0.0cm},clip]{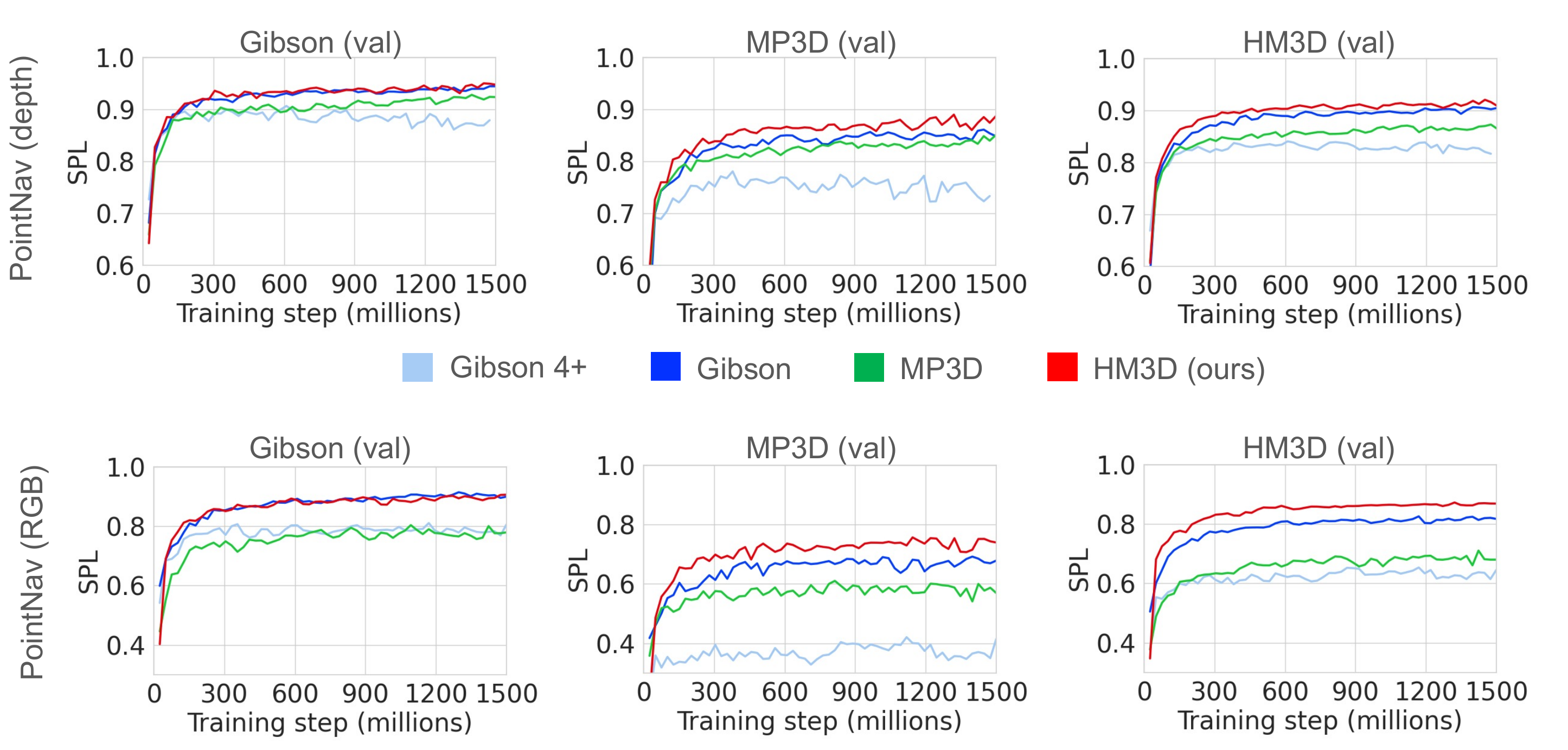}
    \captionof{figure}{\small \textbf{PointNav validation performance vs. training steps:} The top row shows results with depth-only agents, and the bottom row shows results with RGB-only agents.
    Training on HM3D leads to faster convergence and better generalization to newer scenes and datasets.}
    \label{fig:pointnav_curves}
\end{minipage}

\begin{minipage}{\linewidth}
    \centering
    \resizebox{0.8\textwidth}{!}{%
    \begin{tabular}{@{}cl|cc|cc|cc@{}}
    \toprule
    &                         & \multicolumn{2}{c|}{Gibson (test)} & \multicolumn{2}{c|}{MP3D (test)} & \multicolumn{2}{c }{HM3D (test)} \\
    & Dataset                 & Success $\uparrow$ &  SPL $\uparrow$  &Success $\uparrow$ & SPL $\uparrow$   & Success $\uparrow$ & SPL $\uparrow$    \\ \hline
    \parbox[t]{2mm}{\multirow{4}{*}{\rotatebox[origin=c]{90}{Depth}}} 
    & MP3D                    & \ms{ 0.97}{ 0.01} & \ms{ 0.90}{ 0.01} & \ms{ 0.89}{ 0.00} & \ms{ 0.80}{ 0.00} & \ms{ 0.96}{ 0.00} & \ms{ 0.87}{ 0.00} \\
    & Gibson 4+               & \ms{ 0.96}{ 0.02} & \ms{ 0.90}{ 0.02} & \ms{ 0.77}{ 0.01} & \ms{ 0.68}{ 0.01} & \ms{ 0.93}{ 0.00} & \ms{ 0.84}{ 0.00} \\
    & Gibson                  &\bms{ 1.00}{ 0.00} &\bms{ 0.94}{ 0.01} & \ms{ 0.90}{ 0.00} & \ms{ 0.80}{ 0.00} & \ms{ 0.98}{ 0.00} & \ms{ 0.90}{ 0.00} \\
    & HM3D                    &\bms{ 1.00}{ 0.00} &\bms{ 0.93}{ 0.01} &\bms{ 0.94}{ 0.00} &\bms{ 0.83}{ 0.00} &\bms{ 0.99}{ 0.00} &\bms{ 0.92}{ 0.00} \\ \midrule
    \parbox[t]{2mm}{\multirow{4}{*}{\rotatebox[origin=c]{90}{RGB}}} 
    & MP3D                    & \ms{ 0.91}{ 0.01} & \ms{ 0.73}{ 0.03} & \ms{ 0.72}{ 0.01} & \ms{ 0.56}{ 0.01} & \ms{ 0.85}{ 0.00} & \ms{ 0.68}{ 0.00} \\
    & Gibson 4+               & \ms{ 0.88}{ 0.01} & \ms{ 0.73}{ 0.03} & \ms{ 0.44}{ 0.00} & \ms{ 0.35}{ 0.00} & \ms{ 0.77}{ 0.00} & \ms{ 0.62}{ 0.00} \\
    & Gibson                  & \ms{ 0.96}{ 0.00} & \ms{ 0.87}{ 0.01} & \ms{ 0.82}{ 0.01} & \ms{ 0.68}{ 0.01} & \ms{ 0.94}{ 0.00} & \ms{ 0.82}{ 0.00} \\
    & HM3D                    &\bms{ 1.00}{ 0.01} &\bms{ 0.90}{ 0.02} &\bms{ 0.85}{ 0.01} &\bms{ 0.71}{ 0.01} &\bms{ 0.98}{ 0.00} &\bms{ 0.87}{ 0.00} \\ \bottomrule
    \end{tabular}
    }
    \caption{\small \textbf{PointNav test performance} on multiple navigation metrics. We report the mean and standard deviation by training on 1 random seed, and evaluating on 3 random seeds. The $1^{st}$ column indicates whether the agent uses depth or RGB inputs. The HM3D agents reach $100\%$ navigation success for both sensors on Gibson (test). In the majority of cases, HM3D agents significantly outperform the other agents on both metrics. Thus, training on HM3D greatly benefits embodied agents.\vspace{-0.2cm}}
    \label{tab:pointnav_table} 
\end{minipage}
\end{table*}

\subsection{Experimental setup}

We benchmark embodied agents on the PointGoal navigation (a.k.a.~PointNav) task~\cite{anderson2018evaluation}, which has served as a standard testbed for exploring ideas in navigation~\cite{wijmans2019dd,perez2020robot,ramakrishnan2020occupancy,karkus2021differentiable} and a starting point for more semantic tasks~\cite{batra2020objectnav,ku2020room,anderson2018vision}.
In PointNav, an agent is randomly spawned inside a novel environment, and is given a navigation goal coordinate $(\Delta x, \Delta y)$ relative to its starting location.
It has to efficiently navigate to this goal using visual inputs (RGB or depth sensors).
Specifically, we consider the PointNav-v1 task~\cite{savva2019habitat}, where the agent's observation space consists of $256 \times 256$ RGB-D visual inputs, and GPS+Compass readings for localization. 
The agent's action space is [\code{MOVE FORWARD}, \code{TURN LEFT}, \code{TURN RIGHT}, \code{STOP}]. The forward step size is $0.25\si{m}$ and the turn angle is $10^{\circ}$. 
The agent succeeds if it reaches within $0.2\si{m}$ of the goal location and executes \code{STOP}. 
We evaluate PointNav performance using (1) Success, which measures the fraction of episodes successfully completed, and (2) SPL, which measures the efficiency of navigation relative to the shortest paths~\cite{anderson2018evaluation}.

We train and evaluate PointNav agents on Gibson 4+, Gibson, MP3D, and HM3D datasets.
We divide the $1@000$ HM3D scenes into disjoint sets of $800$ train / $100$ val / $100$ test scenes.
We use the standard train / val / test splits for Gibson 4+ and MP3D~\cite{savva2019habitat}. 
We create new PointNav episode datasets for the full Gibson train scenes and HM3D using the generation script from~\citet{savva2019habitat}.
Specifically, we generate 4.11M train episodes for Gibson,
and 8.0M train / 2500 val / 2500 test episodes for HM3D.\footnote{$10@000$ episodes per train scene.  25 episodes per val/test scene.} These splits are publicly available to aid reproducibility: {\small\url{https://github.com/facebookresearch/habitat-lab}}.
In general, MP3D has the hardest episodes and Gibson has the easiest episodes.
See Appendix~\ref{suppsec:pointnav_dataset} for a comparison between the different PointNav episode datasets.

We use a standard agent architecture for training on different datasets~\cite{wijmans2019dd}. 
A ResNet-50 backbone extracts visual features~\cite{he2016deep}, 
and an MLP extracts location features from GPS+compass readings. An LSTM 
state-encoder aggregates these features over time~\cite{hochreiter1997lstm}, and fully-connected layers are used to predict action logits (i.e., the policy) and state values (i.e., the value function). Actions are then stochastically sampled from the predicted action logits. We train the agent using DD-PPO for 1.5 billion frames which was shown to be sufficient to achieve near state-of-the-art performance~\cite{wijmans2019dd}.

We separately benchmark agents for two types of inputs. For `RGB inputs', the agent navigates using RGB and GPS+compass sensors. For `depth inputs', the agent navigates using depth and GPS+compass sensors.
For brevity, `X agent' refers to an agent trained on dataset X (e.g., Gibson agent), and `X agent \res{R}{D}' denotes the SPL performance of X agent with RGB (R) and depth (D) inputs.
\Cref{fig:pointnav_curves} and \Cref{tab:pointnav_table} present results for all agents and datasets. We analyze these results next to answer 3 key questions. 

\texthead{1) Is HM3D beneficial for training PointNav agents?} Consider the validation curves in \Cref{fig:pointnav_curves}. Both HM3D agents (RGB and depth inputs) converge faster and perform better than corresponding agents trained on other datasets.
Specifically, the HM3D agent closely follows the Gibson agent on Gibson (val) and outperforms it on MP3D (val) and HM3D (val).
The validation performance of the HM3D agent rapidly outpaces the MP3D and Gibson 4+ agents on all cases.
The test performance in \Cref{tab:pointnav_table} confirms the above trends. 
On Gibson (test) with depth inputs, the HM3D agent matches the Gibson agent achieving 0.93 SPL and 1.0 success. On all other cases, the HM3D agents outperform the other agents by a large margin. For example, on MP3D (test), the HM3D agent \res{0.71}{0.83} significantly outperforms the second-best Gibson agent \res{0.68}{0.80}. Thus, HM3D is pareto-optimal since the HM3D agents achieve the best performance on all test sets. 

\texthead{2) Are HM3D scenes diverse in terms of visual appearance and 3D layouts?}
Diversity in visual appearance and 3D layouts in the training scenes is essential for generalization to 
novel scenes and datasets, and adaptability to difficult PointNav episodes.
From Table~\ref{tab:pointnav_table}, HM3D agents (both RGB and depth) outperform the next best method on MP3D (test) by 3 SPL points,
and achieve perfect success on Gibson (test). This is impressive generalization since the
HM3D agent had not observed any Gibson or MP3D scenes during training, and yet was able to overcome the domain gap in
appearance and layouts of the scenes. This attests to the visual richness and layout diversity of HM3D which enables good generalization to \emph{previously unseen scenes and datasets}. \vspace{-0.1cm}

\begin{figure}
    \centering
    \includegraphics[width=0.95\textwidth,trim={0 5.7cm 0.5cm 0},clip]{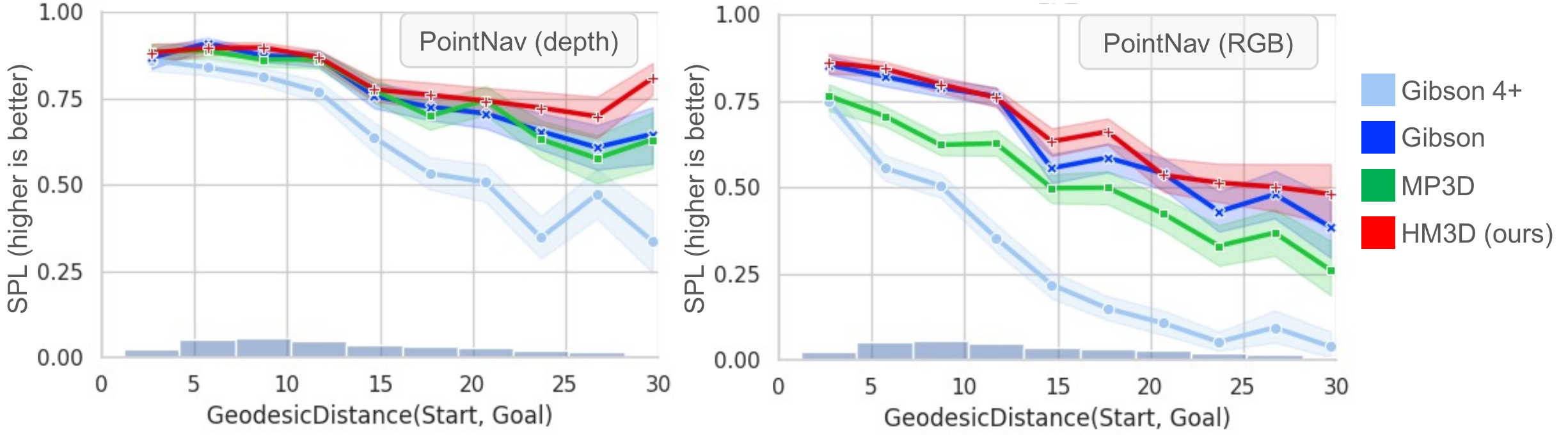}
    \caption{\small \textbf{PointNav SPL vs. episode difficulty:} We divide the MP3D (test) episodes into bins based on the geodesic distance between the start and goal positions (a measure of episode difficulty). For each bin, we report the mean and stddev SPL for PointNav agents with depth (top figure) and RGB (bottom figure) inputs. The diversity and complexity of HM3D layouts allows the HM3D agents to generalize better to harder episodes. \vspace{-0.3cm}}
    \label{fig:pointnav-histogram}
\end{figure}

\begin{figure}
    \centering
    \includegraphics[width=0.85\textwidth,trim={0 8.5cm 10.0cm 0},clip]{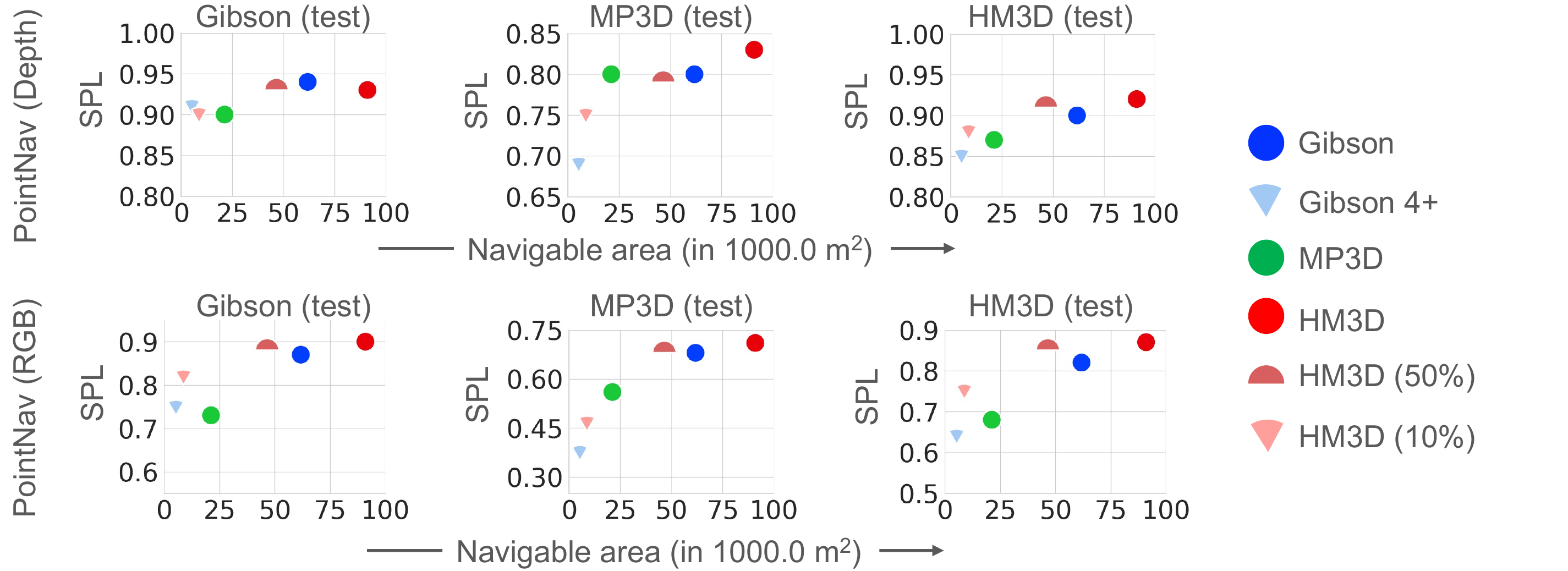}
    \captionof{figure}{\small \textbf{PointNav test performance vs. navigable area:} PointNav performance scales nearly linearly as a function of the total navigable area in training scans.\vspace{-0.3cm}}
    \label{fig:pointnav-size}
\end{figure}

Next, we compare the performance of different agents as a function of the episode difficulty in \Cref{fig:pointnav-histogram}.
We quantify episode difficulty using the geodesic distance between the start and goal locations~\cite{wijmans2019dd}.
We group the MP3D (test) episodes into different bins based on the above metric, and plot the mean and standard deviation of an agent's performance on all episodes in each bin.
We select MP3D (test) since it has highest diversity of difficulty levels (see Appendix~\ref{suppsec:pointnav_dataset}).
We observe that the HM3D agent adapts much better compared to other agents as the episode difficulty increases.
This is yet another indicator that the layouts in HM3D are \emph{complex and diverse}.\vspace{0.2cm}\\
\texthead{3) Does PointNav benefit from scaling up 3D datasets?}
Prior work has verified the data scaling hypothesis for passive perception, i.e., scaling up the dataset can significantly improve performance on various passive perception problems~\cite{deng2009imagenet,lin2014microsoft,sun2017revisiting,yalniz2019billion}. However, this is not well-established in the embodied perception literature due to the lack of large-scale 3D datasets with high quality.
As discussed in earlier sections, HM3D offers large-scale, high visual fidelity, and high-quality reconstructions.
Thus, we use HM3D to test the data scaling hypothesis for embodied AI. 
Specifically, we test the relationship between the training dataset size and the PointNav performance.
For this purpose, we additionally train agents on two random subsets of HM3D containing $10\%$ and $50\%$ of the
scans in the HM3D train split (i.e., 80 and 400 scans respectively). We refer to these agents as HM3D ($10\%$) and HM3D ($50\%$).
\Cref{fig:pointnav-size} shows the PointNav SPL on the test splits as a function of the total navigable area in the training scenes.
We observe that the navigation performance is strongly correlated with the total navigation area (Pearson coefficient $\rho = 0.88$), and that the performance scales near-linearly as the total navigable area increases.
This result is also helpful to decide the data budget for training PointNav agents. Using more data leads to better performance (particularly on the harder episodes in MP3D), but requires more computational resources and time. Depending on the task difficulty and availability of computational resources, researchers can choose the appropriate dataset(s) for experimentation. 
\vspace{0.2cm}\\

\section{Conclusion}

We presented the Habitat-Matterport 3D (HM3D) dataset consisting of $1@000$ building-scale reconstructions from the real world.
To our knowledge, HM3D offers the largest dataset of high-quality 3D reconstructions of interiors for academic research.
Through a series of quantitative analyses we showed that HM3D improves upon existing 3D reconstruction datasets in three ways: significantly larger spatial scale, improved reconstruction completeness, and higher visual fidelity.
We also carried out experiments with PointGoal navigation for embodied AI agents to show that agents trained on HM3D match or outperform agents trained on other datasets even when evaluated on other datasets.
This demonstrates the value of HM3D as a dataset for embodied AI.
Extension of HM3D with object semantics and physical attributes in future work will enable even more embodied AI tasks such as ObjectGoal navigation and object rearrangement.
We hope that HM3D will catalyze research in the area of embodied AI.

{\small
\bibliographystyle{unsrtnat}
\setlength{\bibsep}{0pt}
\bibliography{main}
}

\section{Acknowledgements}
We thank all the volunteers who contributed to the dataset curation effort:
Harsh Agrawal,
Sashank Gondala,
Rishabh Jain,
Shawn Jiang,
Yash Kant,
Noah Maestre,
Yongsen Mao,
Abhinav Moudgil,
Sonia Raychaudhuri,
Ayush Shrivastava,
Andrew Szot,
Joanne Truong,
Madhawa Vidanapathirana,
Joel Ye.
We thank our collaborators at Matterport for their contributions to the dataset:
Conway Chen,
Victor Schwartz,
Nicole Rogers, 
Sachal Dhillon,
Raghu Munaswamy,
Mark Anderson.

\section{Licenses for referenced datasets}
Gibson: {\small\url{http://svl.stanford.edu/gibson2/assets/GDS_agreement.pdf}}\\
Matterport3D: {\small\url{http://kaldir.vc.in.tum.de/matterport/MP_TOS.pdf}} \\
ScanNet: \small{\url{http://kaldir.vc.in.tum.de/scannet/ScanNet_TOS.pdf}} \\
Replica: \small{\url{https://github.com/facebookresearch/Replica-Dataset/blob/master/LICENSE}}
\vfill
\pagebreak

\appendix
\appendixpage
\addappheadtotoc

\setcounter{section}{0}
\setcounter{figure}{0}
\setcounter{table}{0}
\renewcommand{\thesection}{A\arabic{section}}
\renewcommand{\thetable}{A\arabic{table}}
\renewcommand{\thefigure}{A\arabic{figure}}
\section{Limitations of Habitat-Matterport 3D}
\label{suppsec:limitations}

\xhdr{Data acquisition:} The dataset is currently limited to scans from 38 countries. The dataset is restricted to contain data from building-owners who can afford to purchase the Matterport Pro2 sensor (which costs $\sim 3@000\$$) and have internet access to upload data to the cloud. The dataset also excludes regions where the Matterport Pro2 is not available to purchase. Due to these factors, we are limited in the types of regions and neighborhoods which can be included in the dataset. This can introduce an unintended bias into the algorithms developed based on our dataset, where the algorithms work only in a subset of buildings that we encounter in the real world. Nevertheless, this dataset is a significant leap from past building-scale datasets~\cite{xiazamirhe2018gibsonenv,chang2017matterport3d,armeni20163d} that were restricted to labs, residences, and offices. We hope to expand our data set in the future to include scans from many more diverse backgrounds and countries.

\xhdr{Task support:} The dataset only supports geometric tasks in static (i.e., unchanging) environments and does not include semantic annotations. We plan to investigate augmenting the dataset with semantic annotations to tackle high-level understanding tasks like object retrieval. We also plan to study dynamic and changing environments so that our simulations will be fluid rather than static. This would bring simulated training environments closer to the real world, where people and pets freely move around and where everyday objects such as mobile phones, wallets, and shoes are not always in the same spot throughout the day.
\section{Hyperparameters for PointNav experiments}
\label{suppsec:hyperparams}

We use the publicly available implementation of DD-PPO~\cite{wijmans2019dd} from \href{https://github.com/facebookresearch/habitat-lab}{Habitat Lab}. We use the same hyperparameters as~\citet{wijmans2019dd} for our experiments. We use a ResNet-50~\cite{he2016deep} backbone and an LSTM with 512-D hidden states and 2 layers. Following~\citet{wijmans2019dd}, we replace BatchNorm with GroupNorm layers in the ResNet-50 backbone. We use a PPO clip parameter of $0.2$, 2 PPO epochs, 2 mini-batches, a value loss coefficient of $0.5$, entropy coefficient of $0.01$ and a learning rate of $0.00025$. Please see the default configuration \href{https://github.com/facebookresearch/habitat-lab/blob/master/habitat_baselines/config/default.py}{here} for more details. We train each model for 1.5 billion steps (sufficient for convergence) with 256 parallel environments divided between 8 nodes, 4 workers (i.e., 4 GPUs) per node and 8 environments per worker.
\section{Computational requirements}
\label{suppsec:compute}

The PointNav experiments were the most computationally expensive of all our experiments. Each experiment is run in our internal cluster in a distributed fashion over 8 nodes, with 4 GPUs per node. Each GPU (32 in total) is a Volta 16/32 GB. Training an agent takes 2-3 days with depth inputs, and 4-5 days with RGB inputs.
\section{Accessing Habitat-Matterport 3D dataset}
\label{suppsec:accessing}

HM3D is free and available for academic, non-commercial research here: \\
{\small\url{https://matterport.com/habitat-matterport-3d-research-dataset}} \\
The terms of use are available here: \\
{\small\url{https://matterport.com/matterport-end-user-license-agreement-academic-use-model-data}}

\section{Habitat-Matterport 3D dataset collection process}
\label{suppsec:dataset}

The 1000 scans in HM3D were collected by \href{https://matterport.com/}{Matterport Inc.}~in collaboration with the \href{https://aihabitat.org/}{Habitat team} at Facebook AI Research. Matterport directly contacted its users explicitly requesting them to contribute their scans for open-sourced Embodied AI research (see mailer snippet below). 

\begin{displayquote}
    Imagine if firefighters could ask a robot to detect where smoke is coming from within your house, then command it to find people who need help. Or, if you could ask an AI assistant to locate your car keys. To realize innovations like these, robots and AI assistants need to be trained in how to act in multiple environments. They must learn to recognize and navigate through 3D spaces. That’s where you come in. We have identified your Matterport 3D model as an ideal space for an open-source AI project focused on furthering such causes.
\end{displayquote}

Each user agreed to the following terms while contributing their scans for the dataset.
\begin{displayquote}
    I agree to allow Matterport to use the Space(s) (including all related imagery) that I have designated in this form for academic and/or non-commercial purposes as further provided in the Matterport Terms and Conditions for Academic and Non-Commercial Use of Spaces, without payment by Matterport for such use. I affirm that I have all necessary rights, consents and permissions relating to my Space(s) necessary to grant the foregoing permission. By checking this box, I specifically agree to all of the provisions of the Matterport Terms and Conditions for Academic and Non-Commercial Use of Spaces $\&$ Matterport Privacy Policy.
\end{displayquote}

\begin{figure}[t]
    \centering
    \includegraphics[width=\linewidth,trim={0 7cm 2cm 0},clip]{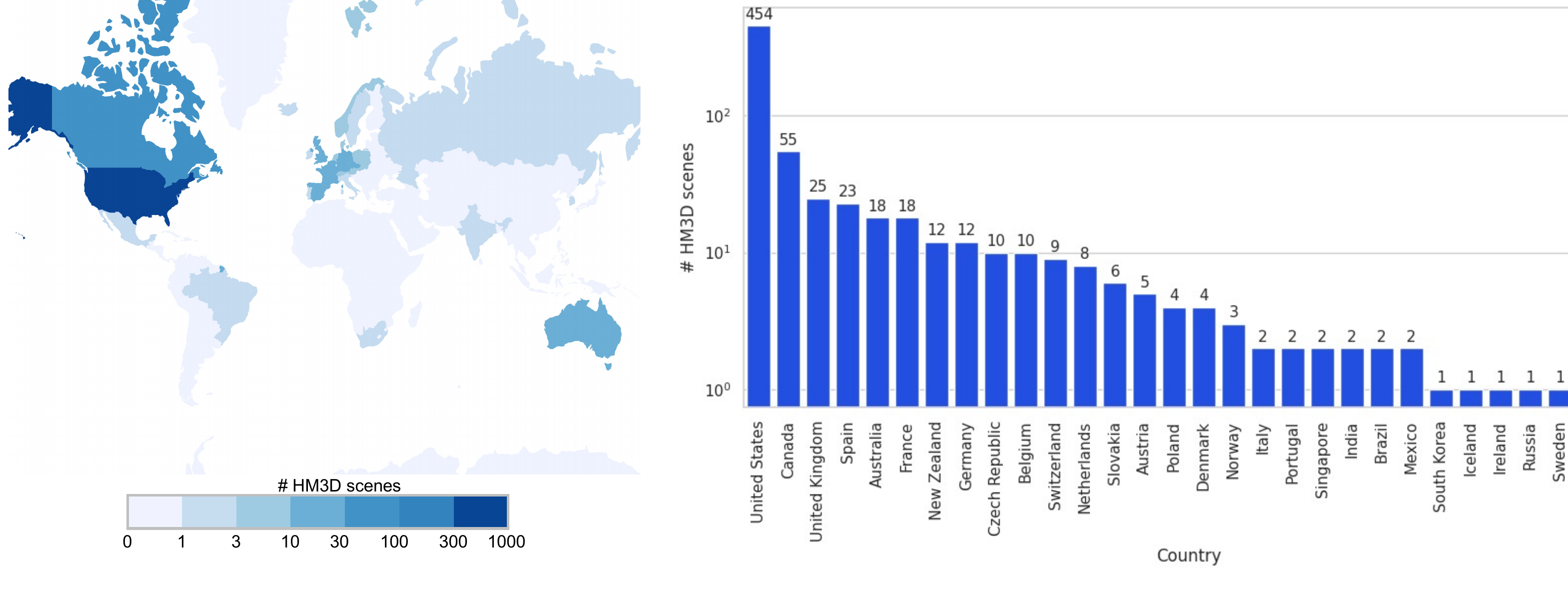}
    \caption{Visualizing the distribution of countries included in HM3D scenes. In total, we obtain scans from users in 38 countries. The left plot shows the distribution over different countries in the world map with the hue representing the number of scenes. The right plot shows a histogram over the different countries. Note: Due to missing data from certain scans, we only visualize the statistics over 30/38 countries in this figure.}
    \label{suppfig:hm3d_regions}
\end{figure}

After obtaining scans from users, Matterport used commercially reasonable efforts to try and obscure personally identifiable information such as pictures of people or faces, names, documents with personal information, diplomas, driver's licenses, email addresses, phone numbers, street addresses, personal notes / letters / envelopes, employer information, license plates, and street names. Human reviewers were asked to preview images from every scanned location to check for the above information and annotated each instance of personal information using a label. Any personal information identified in the previous step was blurred using a pixel-wise blur mask. The blurred data was used to recreate the 3D scans. A helpful FAQ regarding this process can be found here: {\small \url{https://go.matterport.com/ProjectHabitat.html}}. Overall, the dataset contains scans from users in 38 countries (see \Cref{suppfig:hm3d_regions}).

\section{PointNav validation results}
\label{suppsec:pointnav_quantitative}

In Figure 6 from the main paper, we presented the validation performance as a function of the training steps. Now, we present the final validation performance of the best checkpoint (analogous to Table 2 in the main paper). See \Cref{supptab:pointnav_table}. We observe trends similar to the ones observed in the main paper.
The HM3D agent matches the Gibson agent on Gibson (val) with RGB inputs. On all other cases, the HM3D agents outperform the other agents by a good margin, particularly in the RGB case. 

\begin{table*}[t]
    \centering
    \resizebox{0.8\textwidth}{!}{%
    \begin{tabular}{@{}cl|cc|cc|cc@{}}
    \toprule
    &                         & \multicolumn{2}{c|}{Gibson (val)} & \multicolumn{2}{c|}{MP3D (val)} & \multicolumn{2}{c }{HM3D (val)} \\
    & Dataset                 & Success $\uparrow$ &  SPL $\uparrow$  &Success $\uparrow$ & SPL $\uparrow$   & Success $\uparrow$ & SPL $\uparrow$    \\ \hline
    \parbox[t]{2mm}{\multirow{4}{*}{\rotatebox[origin=c]{90}{Depth}}} 
    & MP3D                    & \ms{ 0.98}{ 0.00} & \ms{ 0.92}{ 0.00} & \ms{ 0.93}{ 0.01} & \ms{ 0.84}{ 0.01} & \ms{ 0.96}{ 0.00} & \ms{ 0.87}{ 0.00} \\
    & Gibson 4+               & \ms{ 0.96}{ 0.00} & \ms{ 0.91}{ 0.00} & \ms{ 0.88}{ 0.00} & \ms{ 0.76}{ 0.00} & \ms{ 0.93}{ 0.00} & \ms{ 0.84}{ 0.00} \\
    & Gibson                  & \ms{ 0.99}{ 0.00} & \ms{ 0.94}{ 0.00} & \ms{ 0.95}{ 0.00} & \ms{ 0.86}{ 0.01} & \ms{ 0.98}{ 0.00} & \ms{ 0.90}{ 0.00} \\
    & HM3D                    &\bms{ 1.00}{ 0.00} &\bms{ 0.95}{ 0.00} &\bms{ 0.96}{ 0.00} &\bms{ 0.87}{ 0.00} &\bms{ 0.99}{ 0.00} &\bms{ 0.91}{ 0.00} \\ \midrule
    \parbox[t]{2mm}{\multirow{4}{*}{\rotatebox[origin=c]{90}{RGB}}} 
    & MP3D                    & \ms{ 0.93}{ 0.01} & \ms{ 0.78}{ 0.00} & \ms{ 0.78}{ 0.01} & \ms{ 0.59}{ 0.01} & \ms{ 0.84}{ 0.00} & \ms{ 0.67}{ 0.00} \\
    & Gibson 4+               & \ms{ 0.88}{ 0.00} & \ms{ 0.78}{ 0.01} & \ms{ 0.49}{ 0.00} & \ms{ 0.36}{ 0.00} & \ms{ 0.77}{ 0.01} & \ms{ 0.62}{ 0.00} \\
    & Gibson                  &\bms{ 0.99}{ 0.00} &\bms{ 0.91}{ 0.00} & \ms{ 0.86}{ 0.01} & \ms{ 0.69}{ 0.01} & \ms{ 0.94}{ 0.00} & \ms{ 0.82}{ 0.00} \\
    & HM3D                    &\bms{ 0.99}{ 0.00} &\bms{ 0.91}{ 0.00} &\bms{ 0.92}{ 0.01} &\bms{ 0.74}{ 0.01} &\bms{ 0.97}{ 0.00} &\bms{ 0.87}{ 0.00} \\ \bottomrule
    \end{tabular}
    }
    \caption{\textbf{PointNav val performance} on multiple navigation metrics. We report the mean and standard deviation by training on 1 random seed, and evaluating on 3 random seeds. The $1^{st}$ column indicates whether the agent uses depth or RGB inputs. The HM3D agents reach $100\%$ navigation success for both sensors on Gibson (val). In the majority of cases, HM3D agents significantly outperform the other agents on both metrics. Thus, training on HM3D greatly benefits embodied agents.}
    \label{supptab:pointnav_table} 
\end{table*}
\begin{figure}[t]
    \centering
    \includegraphics[width=0.60\textwidth,trim={0 5.5cm 0 0},clip]{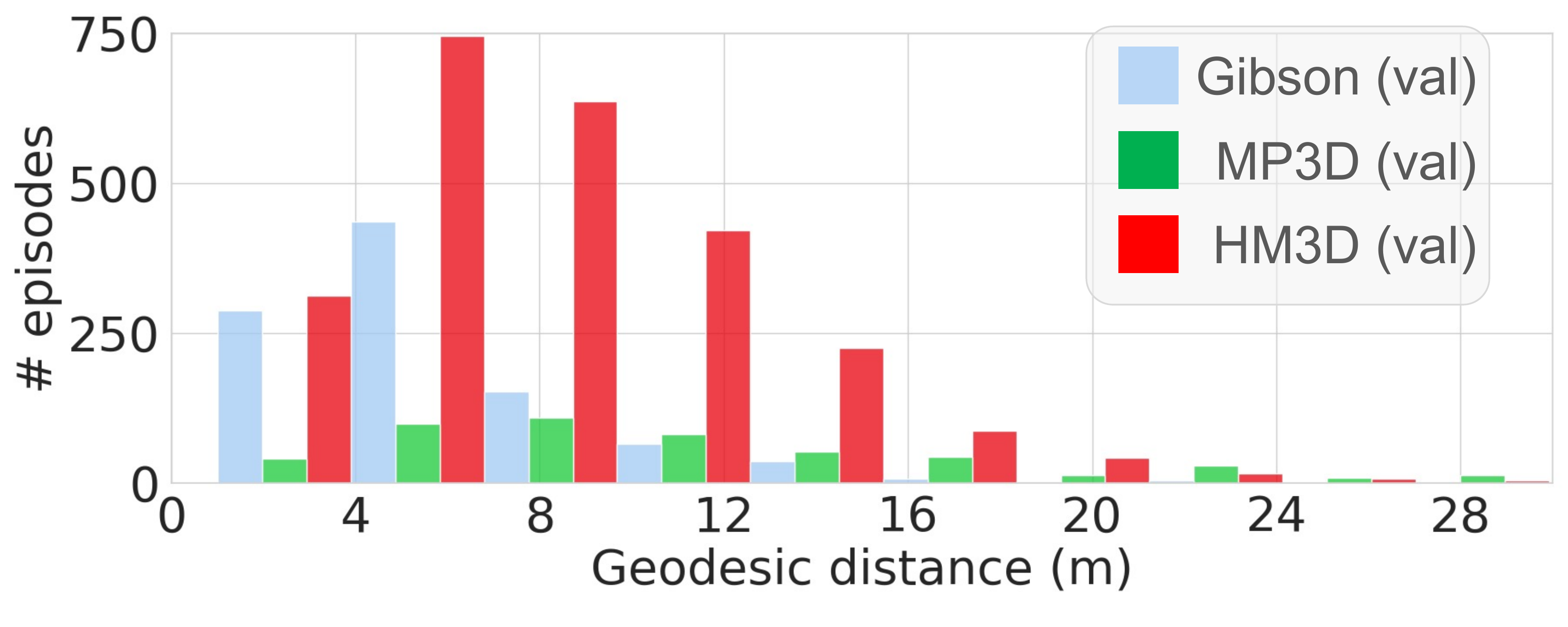}
    \captionof{figure}{ Distribution of PointNav episode difficulties in the val splits of Gibson, MP3D, and HM3D. We group the episodes in each dataset based on the geodesic distance between the start and goal positions. Episodes with larger geodesic distance are generally harder. Gibson has the easiest episodes. MP3D has the hardest episodes.}
    \label{suppfig:pointnav-episodes-difficulty}
\end{figure}

\section{Comparing PointNav episode datasets}
\label{suppsec:pointnav_dataset}

In \Cref{suppfig:pointnav-episodes-difficulty}, we compare the difficulties of the Gibson (val), MP3D (val), and HM3D (val) episode datasets. MP3D has the hardest episodes and Gibson has the easiest episodes.

\section{Dataset characteristics that impact PointNav performance}
\label{suppsec:correlation_study}

In the main paper, we compared datasets along different characteristics such as navigable area, visual fidelity, 3D reconstruction quality, and the utility for training agents for the PointNav task. In Section 4 and Figure 6, we analyzed the impact of dataset size on the PointNav performance, and noted that total navigable area in the training scans are highly correlated with the PointNav results. Here, we perform a complete analysis of how the following factors affect the PointNav performance (on the val splits). 

\texthead{1. EMD (train, val)} measures the dissimilarity between episodes in the train and val splits. We calculate the normalized histogram of geodesic distances between the start and goal locations for each episode in the train and val splits (independently). We then measure the distribution shift between the train and val episodes. This is done by computing the Earth Mover's Distance (EMD) a.k.a Wasserstein Distance between the normalized histograms of geodesic distances for the train and val splits.\\
\texthead{2. KID (mean)} is a measure of visual fidelity of images rendered from each dataset. This is calculated as the mean of KID (Gibson real) and KID (MP3D real) from Table 5(b) in the main paper.\\
\texthead{3. \% defects} is a measure of reconstruction completeness for the 3D scans. For each dataset, this is calculated as the mean of ``\% defects" values from Figure 4 in the main paper. \\
\texthead{4. Navigable area} $(\si{m^2})$ measures the dataset size. It is computed as the overall navigable area in the training scans for each dataset. \\

We compute the above metrics for all the train datasets\footnote{We compute EMD(train, val) for all pairs of train and val sets.}. For a given PointNav val set, we measure the Pearson's correlation between each of the above metrics for a train dataset and the navigation SPL achieved by agents trained on the same dataset (see \Cref{supptab:correlation_study}). As noted in the main paper, we observe that the navigable area is highly correlated with the PointNav performance ($0.82$ to $0.97$) indicating that large-scale datasets are critical for achieving high-quality navigation. For both MP3D (val) and HM3D (val), there is a strong negative correlation of $-0.4$ to $-0.7$ between EMD (train, val) and SPL. This indicates that large distribution shifts between the train and val episodes leads to worse performance. Next, we observe that KID (mean) is weakly correlated with the SPL ($-0.30$ to $0.1$), indicating that visual fidelity may not strongly impact PointNav performance in simulation. In most cases, \% defects is generally uncorrelated with SPL. However, we find a slightly \emph{positive} correlation  ($0.17 \mhp 0.32$) with SPL on MP3D (val). This may be due to the fact that MP3D val scenes have significantly more mesh reconstruction artifacts than Gibson val scenes\footnote{Only Gibson 4+ scenes are used for Gibson (val) and Gibson (test).} (see Figure 4 in main paper). Agents trained on scenes with more mesh reconstruction artifacts adapt better to such testing conditions. Note that these results are not very indicative of the transfer performance to a real robot. It is possible that higher visual fidelity and lower \% defects may be necessary for real-world transfer.

\begin{table}[t]
\centering
\resizebox{1.0\textwidth}{!}{%
\begin{tabular}{@{}l|ccc|ccc@{}}
\toprule
                           & \multicolumn{3}{c|}{PointNav (depth)}  & \multicolumn{3}{c}{PointNav (RGB)}     \\
Factor                     & Gibson (val) & MP3D (val) & HM3D (val) & Gibson (val) & MP3D (val) & HM3D (val) \\ \midrule
EMD (train, val)           & $+0.118$     & $-0.693$   & $-0.412$   & $-0.129$     & $-0.589$   & $-0.584$   \\
KID (mean)                 & $-0.100$     & $+0.016$   & $-0.145$   & $-0.167$     & $-0.062$   & $-0.299$   \\
\% defects                 & $-0.102$     & $+0.302$   & $+0.019$   & $-0.290$     & $+0.168$   & $-0.208$   \\
Navigable area ($\si{m^2}$)& $+0.968$     & $+0.822$   & $+0.900$   & $+0.886$     & $+0.864$   & $+0.884$   \\ \bottomrule
\end{tabular}
}
\vspace{0.1cm}
\caption{We measure the correlation between different dataset characteristics (row) on the PointNav performance (column). EMD (train, val) measures the distribution mismatch between the difficulties of episodes in the train and val PointNav datasets. KID (mean) measures the visual fidelity of images from the train scans. \% defects measures the reconstruction completeness of scans in the train scans. Navigable area measures the total area that is navigable across all the train scans from that dataset.}
\label{supptab:correlation_study}
\end{table}

\section{Example scenes from Habitat-Matterport 3D}
\label{suppsec:examples}

We provide more examples of scenes from Habitat-Matterport 3D in the same style as Figure 2 in the main paper.
In \Cref{suppfig:hm3d_examples_1}, we visualize 5 residences, and in \Cref{suppfig:hm3d_examples_2}, we visualize 5 diverse scenes such as offices, gyms, restaurants, and nightclubs. All 900 scenes from the train and val splits of HM3D can be visualized on the dataset website:  {\small\url{https://aihabitat.org/datasets/hm3d/}}

\begin{table}[t]
\hspace*{-1.5cm}
\resizebox{1.2\textwidth}{!}{
    \centering
    \begin{tabular}{c}
    \includegraphics[width=0.25\textwidth,valign=m]{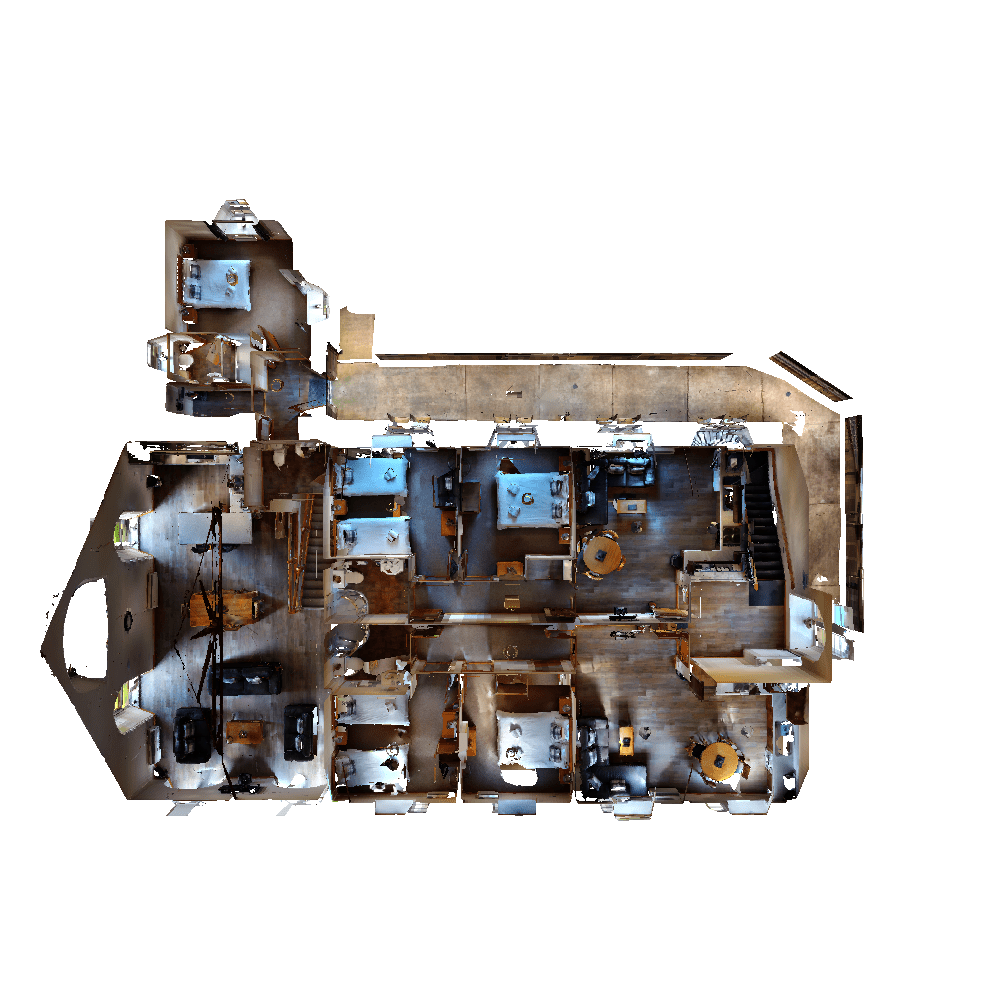}
    \includegraphics[width=0.25\textwidth,valign=m]{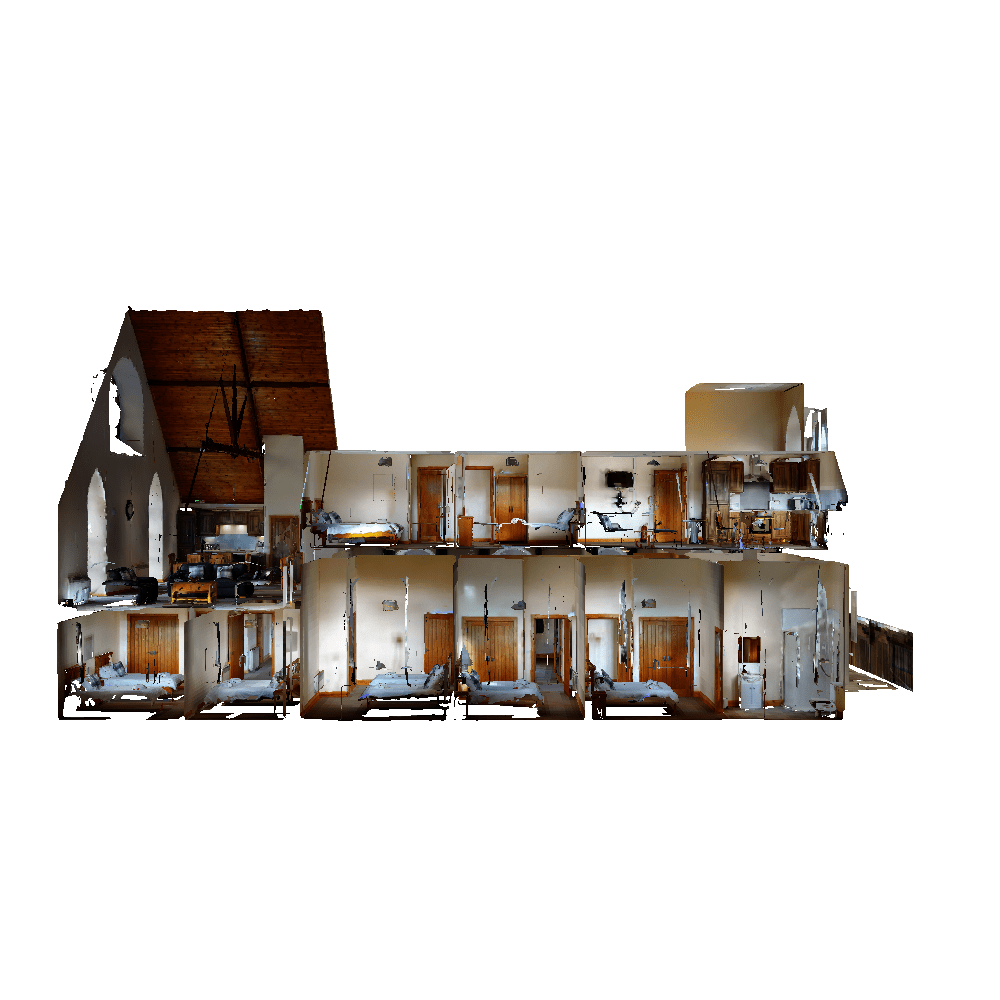}
    \includegraphics[width=0.20\textwidth,valign=m]{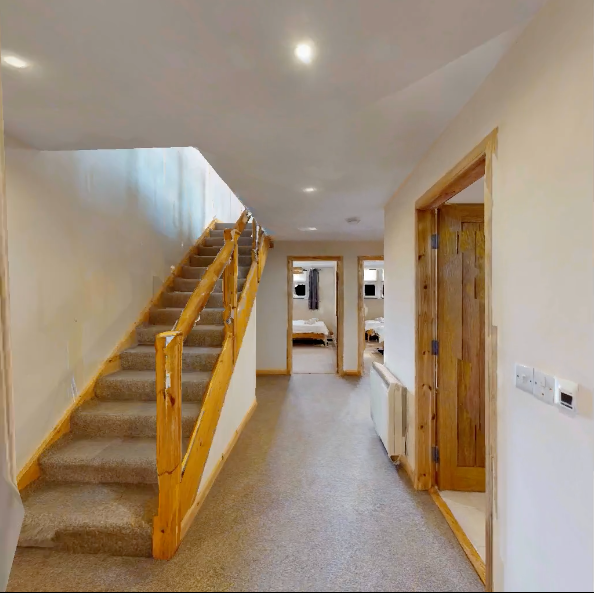}
    \includegraphics[width=0.20\textwidth,valign=m]{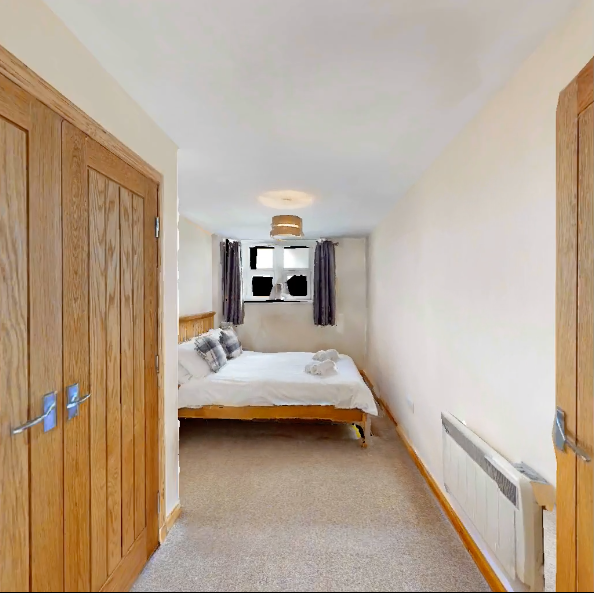} \\
    \includegraphics[width=0.25\textwidth,valign=m]{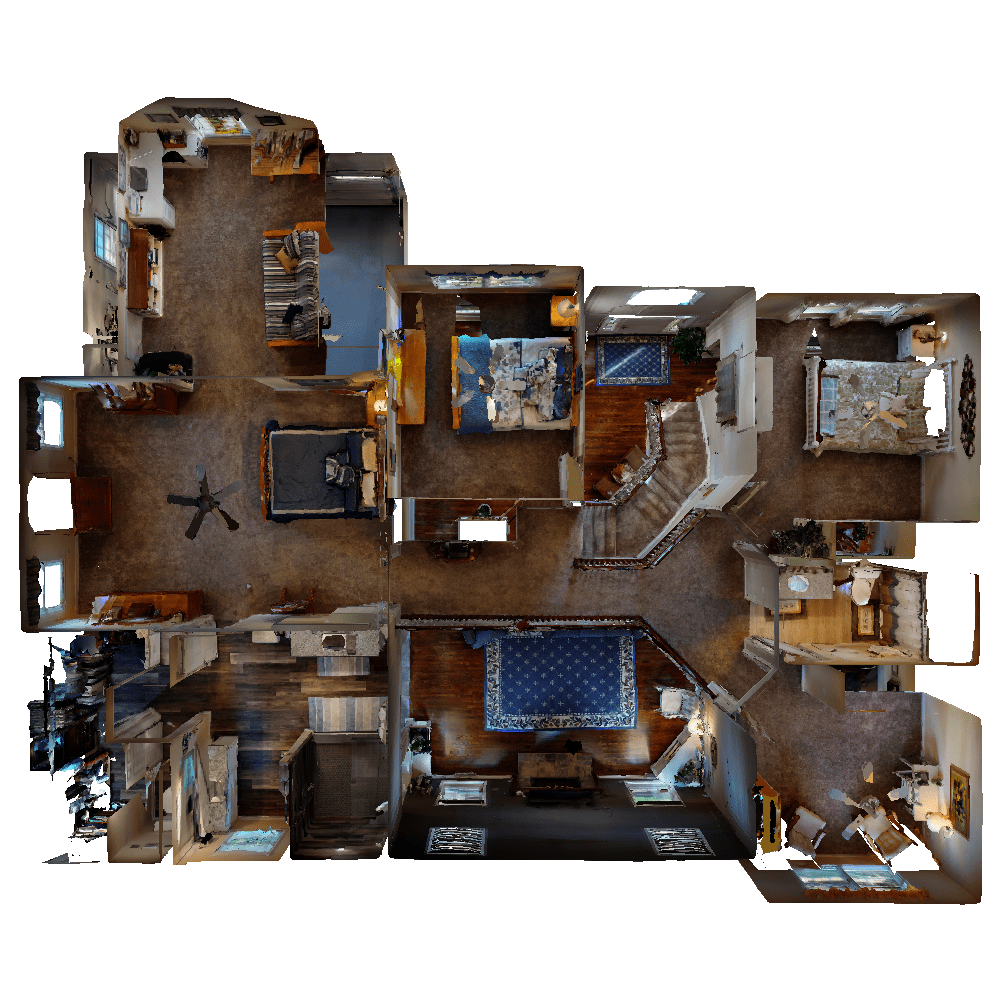}
    \includegraphics[width=0.25\textwidth,valign=m]{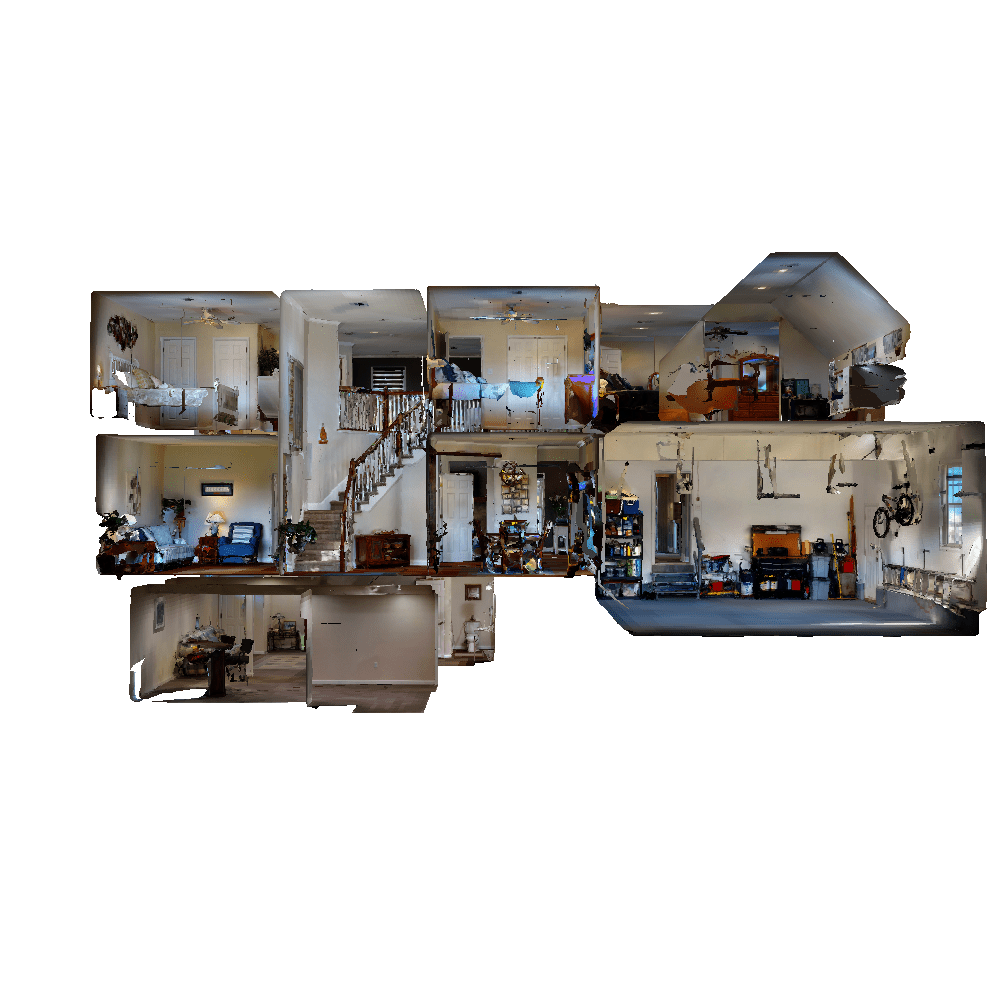}
    \includegraphics[width=0.20\textwidth,valign=m]{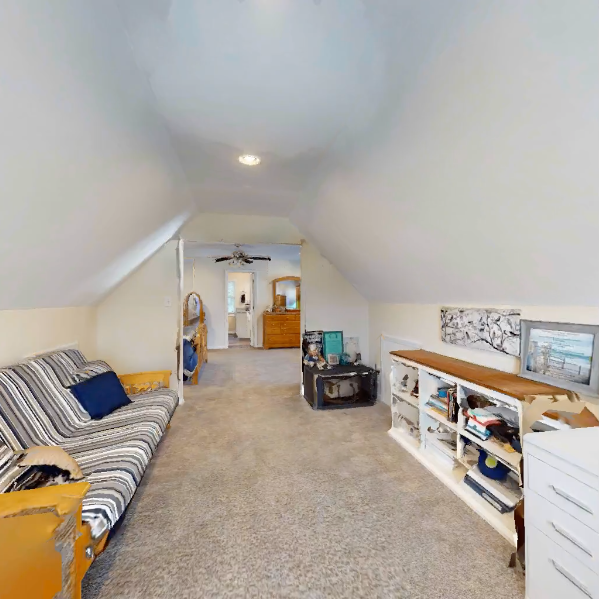}
    \includegraphics[width=0.20\textwidth,valign=m]{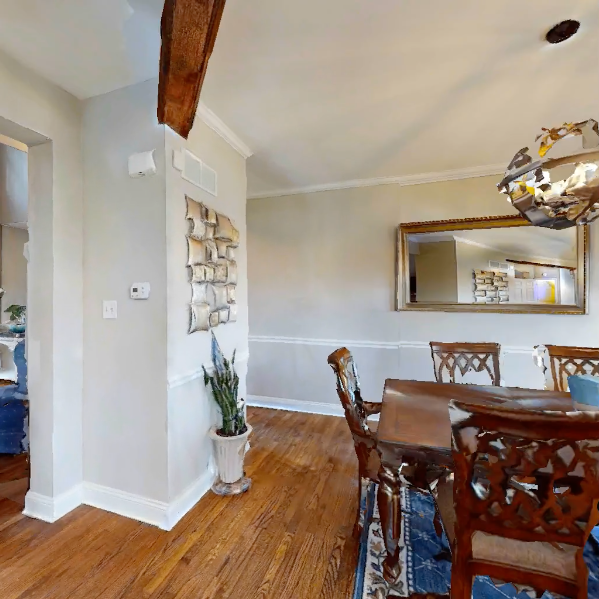} \\
    \includegraphics[width=0.25\textwidth,valign=m]{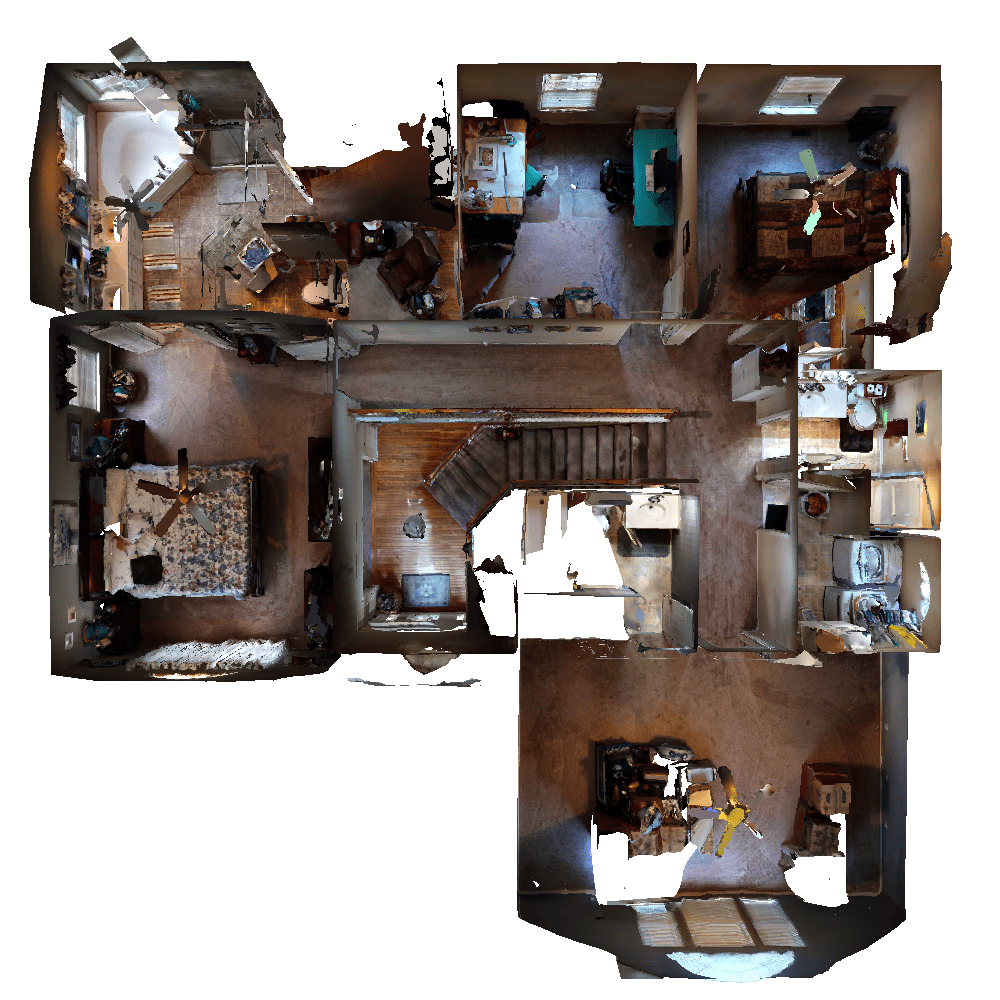}
    \includegraphics[width=0.25\textwidth,valign=m]{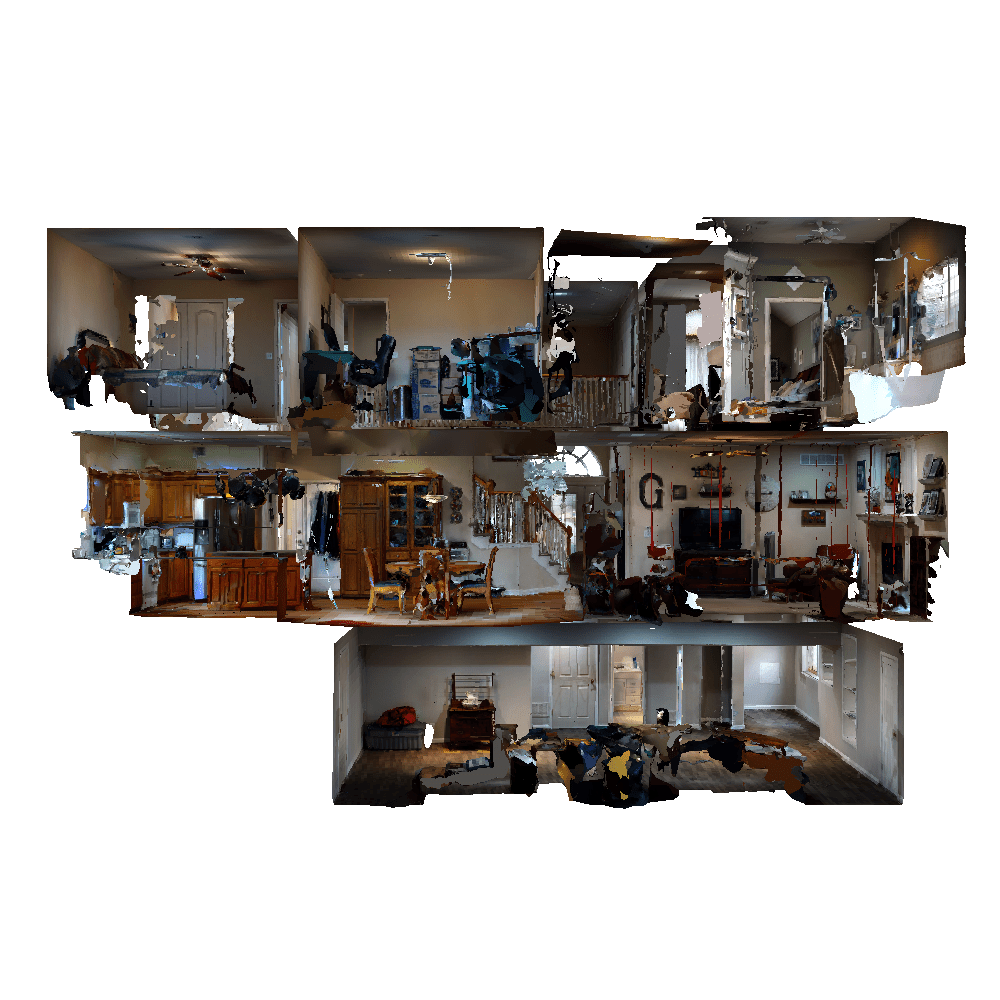}
    \includegraphics[width=0.20\textwidth,valign=m]{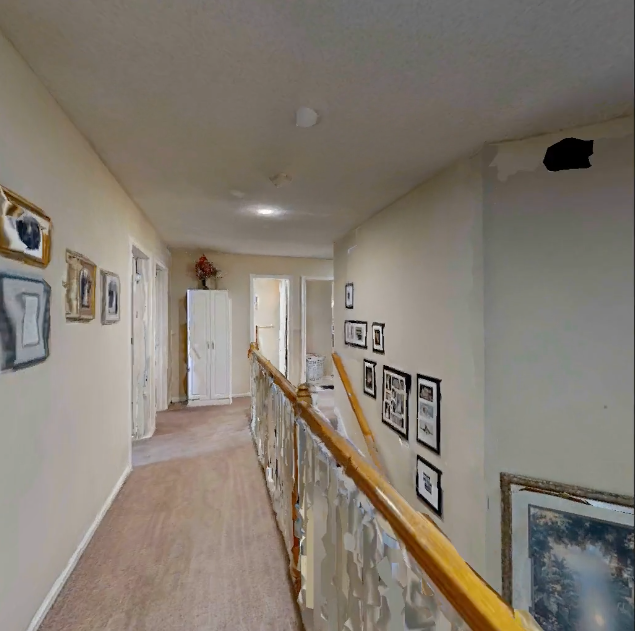}
    \includegraphics[width=0.20\textwidth,valign=m]{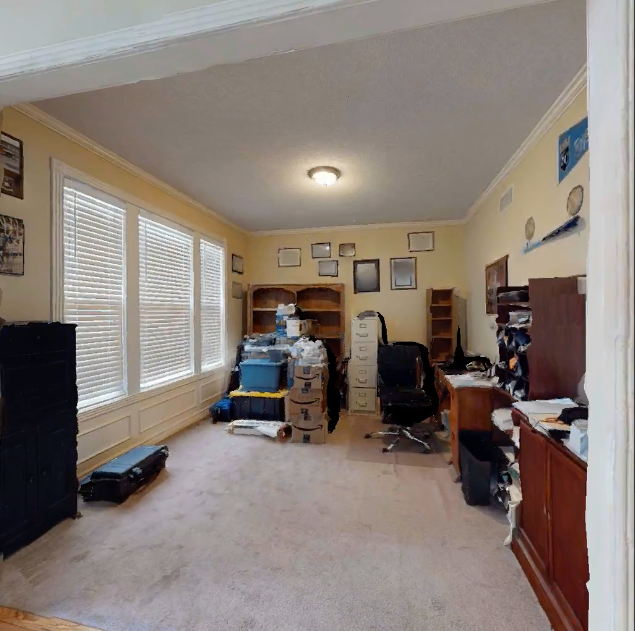} \\
    \includegraphics[width=0.25\textwidth,valign=m]{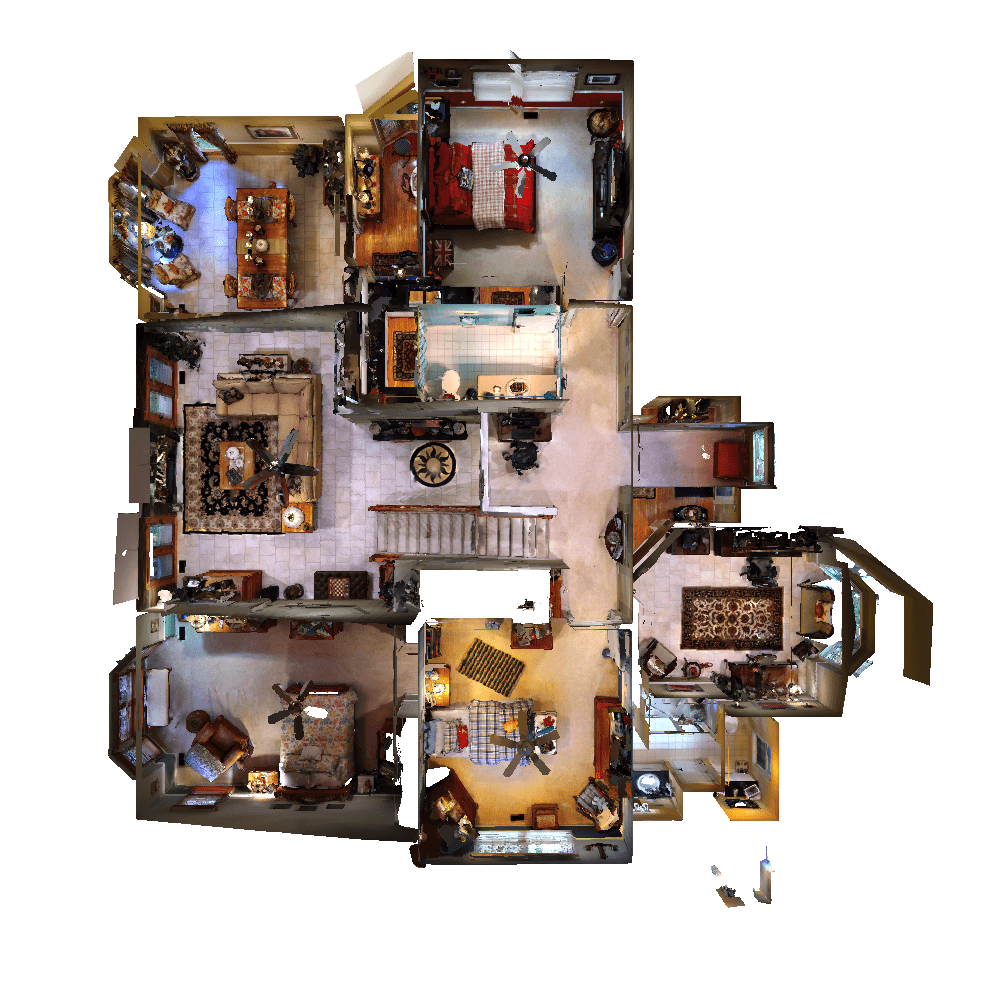}
    \includegraphics[width=0.25\textwidth,valign=m]{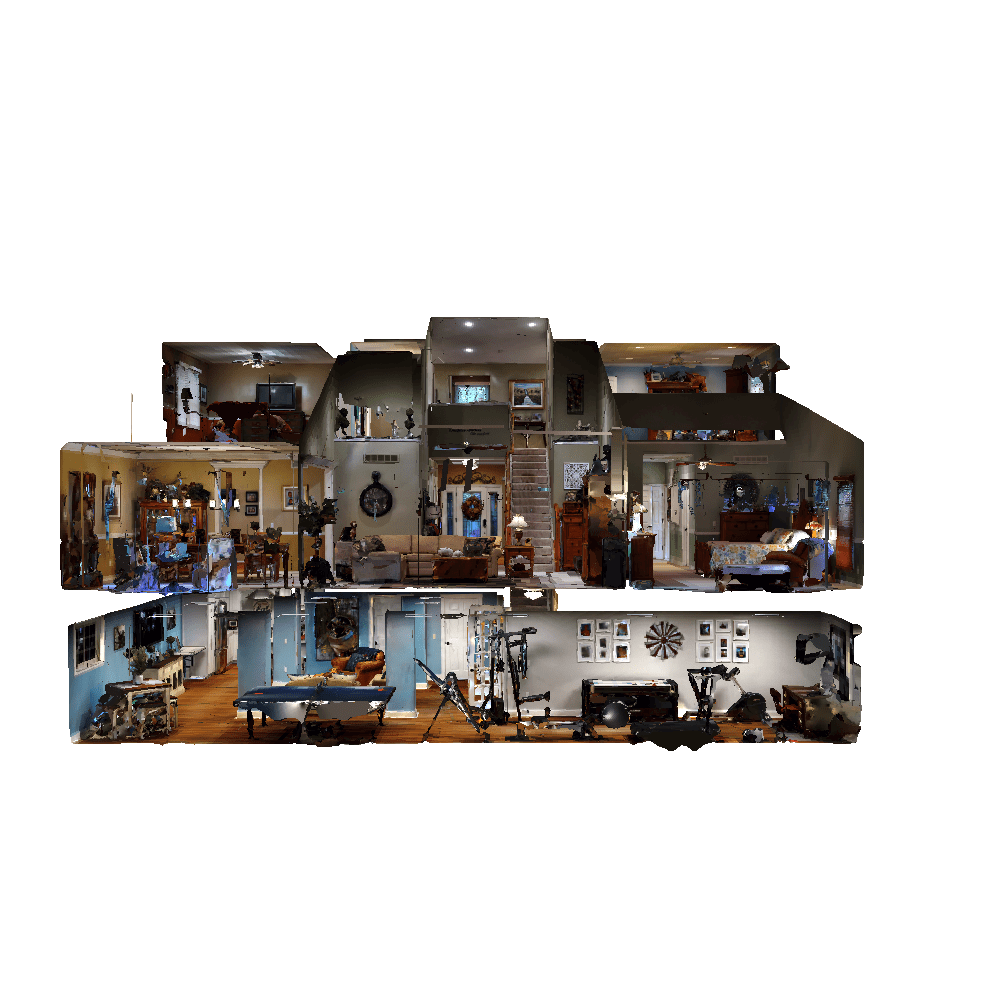}
    \includegraphics[width=0.20\textwidth,valign=m]{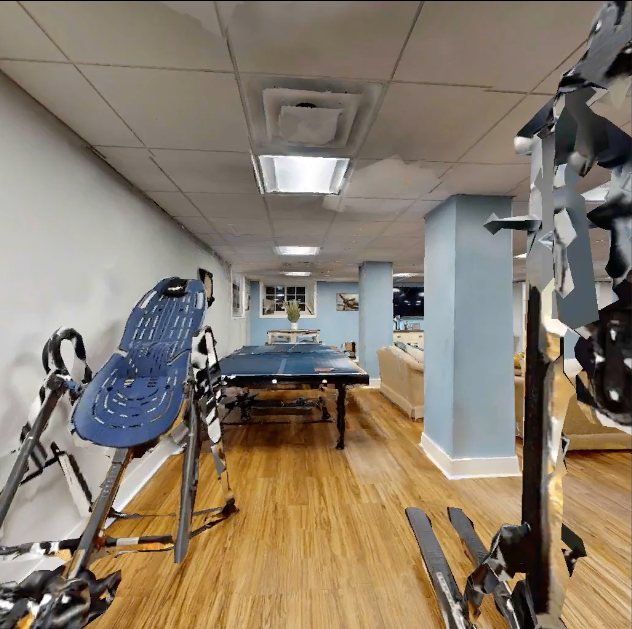}
    \includegraphics[width=0.20\textwidth,valign=m]{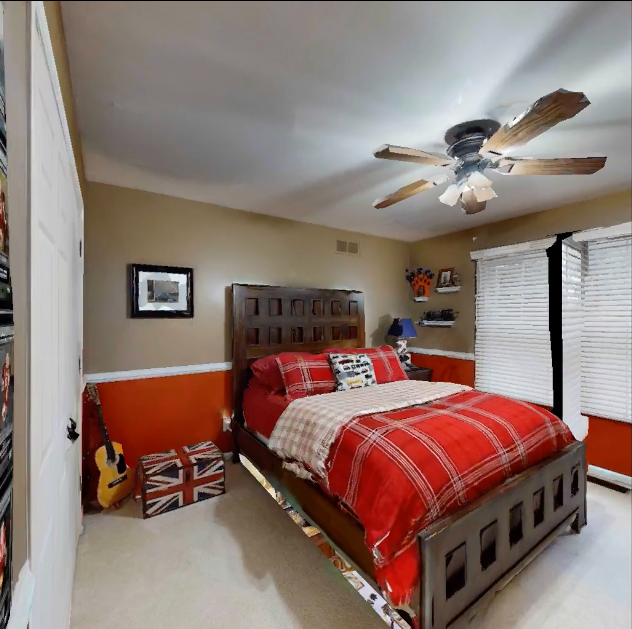} \\
    \includegraphics[width=0.25\textwidth,valign=m]{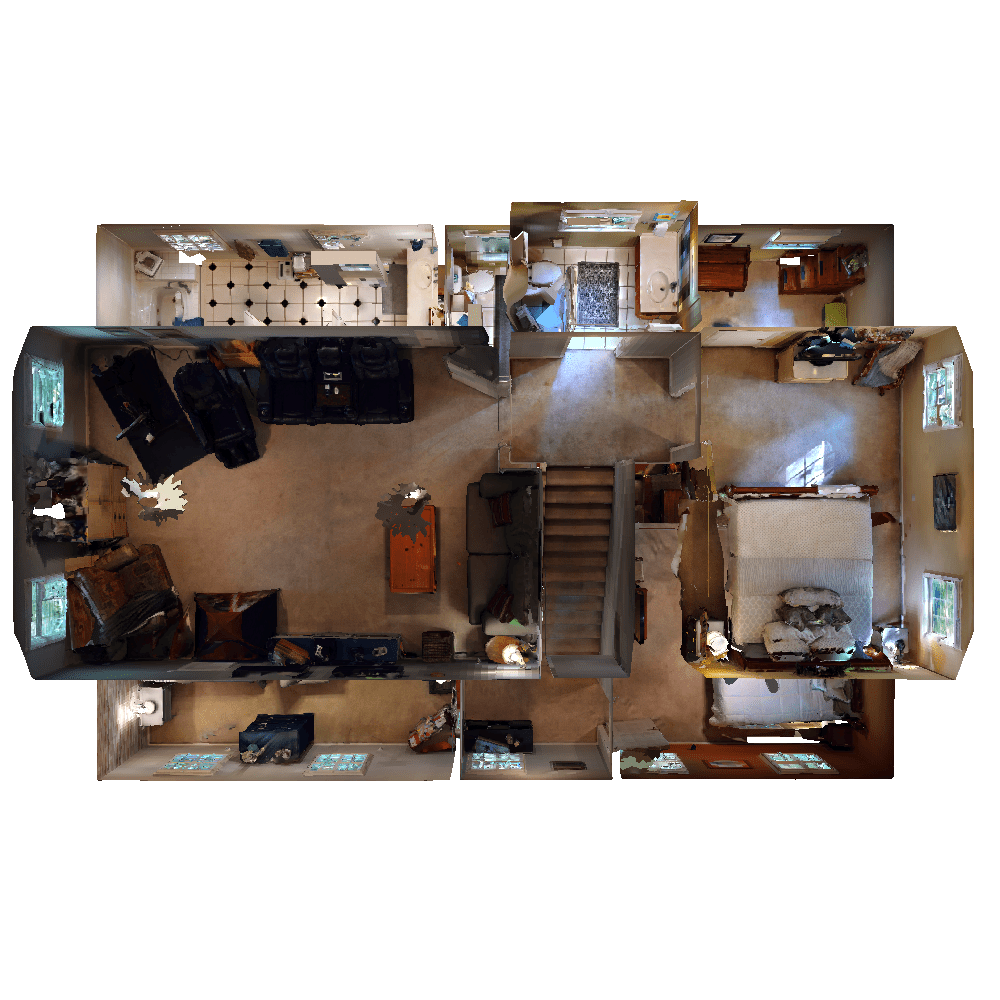}
    \includegraphics[width=0.25\textwidth,valign=m]{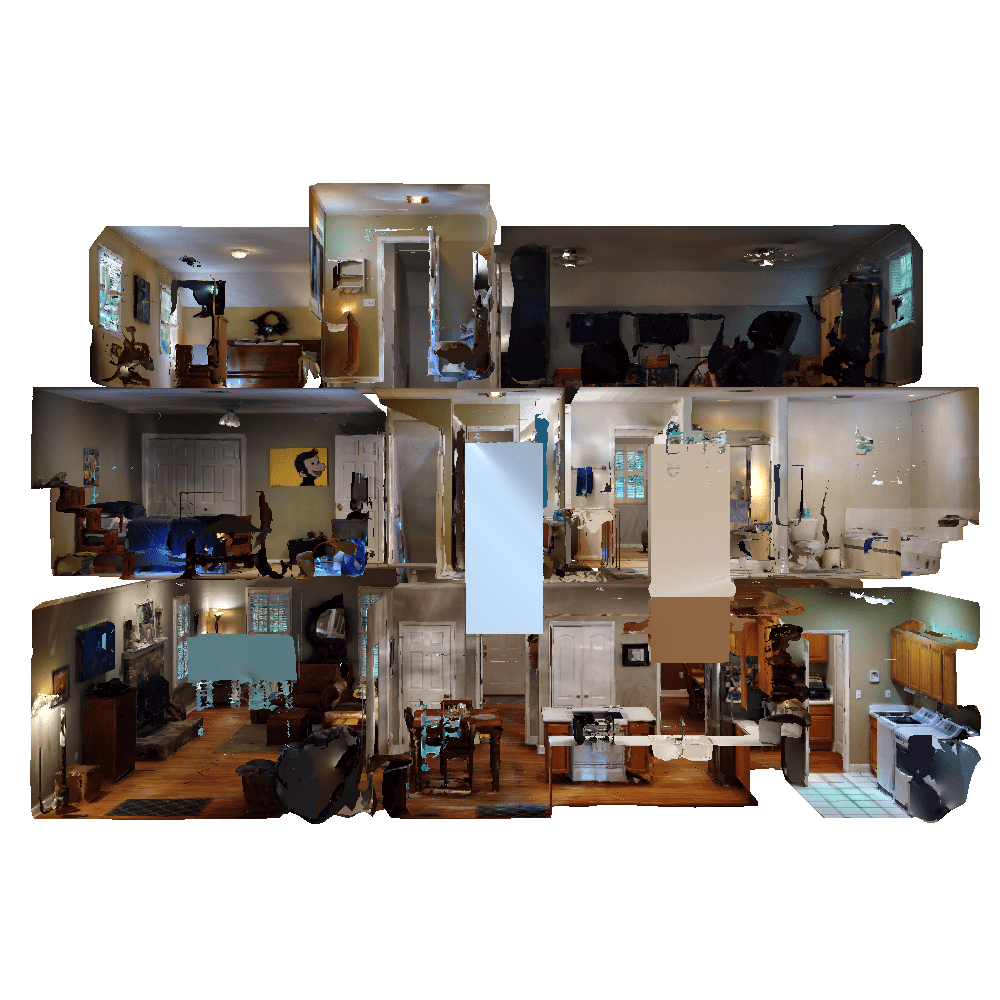}
    \includegraphics[width=0.20\textwidth,valign=m]{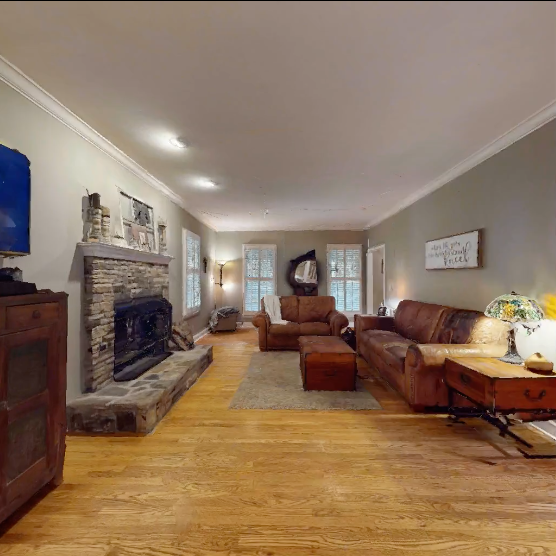}
    \includegraphics[width=0.20\textwidth,valign=m]{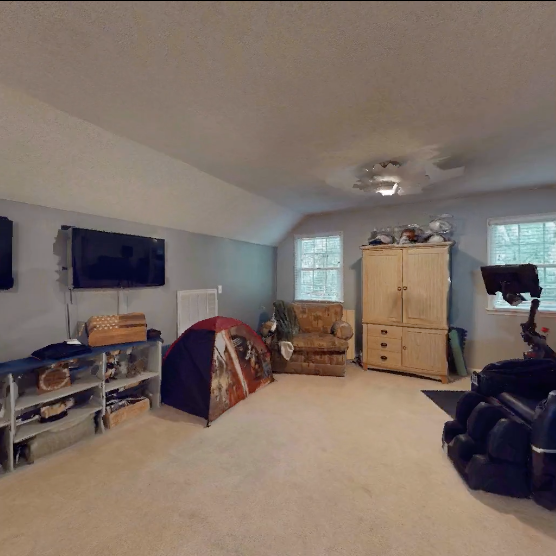} \\
    \end{tabular}
}
\captionof{figure}{Five example residences from the HM3D dataset. From left to right in each row: top-down view, cross section view, and two egocentric views from navigable positions in the scene. The dataset contains multi-floor buildings spanning a wide range of sizes.}
\vspace*{-0.3in}
\label{suppfig:hm3d_examples_1}
\end{table}

\begin{table}[t]
\hspace*{-1.5cm}
\resizebox{1.2\textwidth}{!}{
    \centering
    \begin{tabular}{c}
    \includegraphics[trim={0 0cm 0 0cm},clip,width=0.30\textwidth,valign=m]{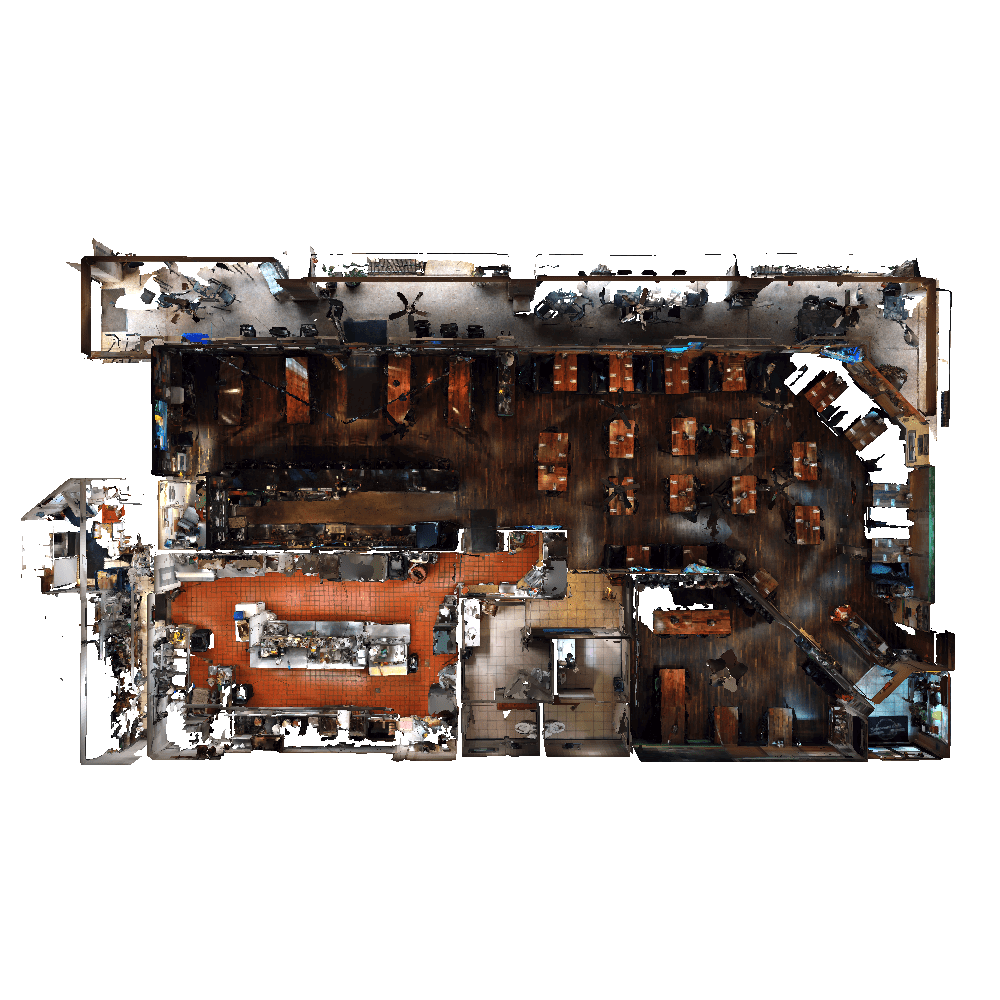}
    \includegraphics[trim={0 3cm 0 3cm},clip,width=0.30\textwidth,valign=m]{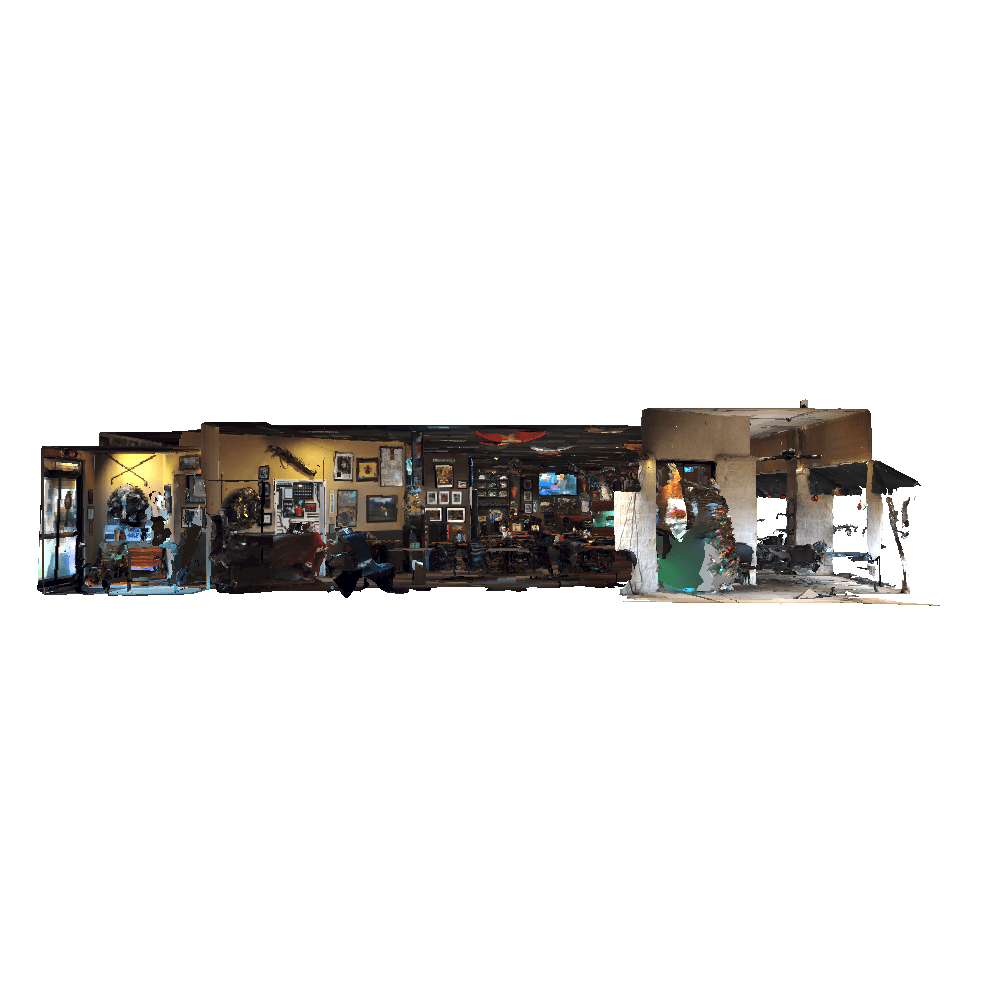}
    \includegraphics[width=0.20\textwidth,valign=m]{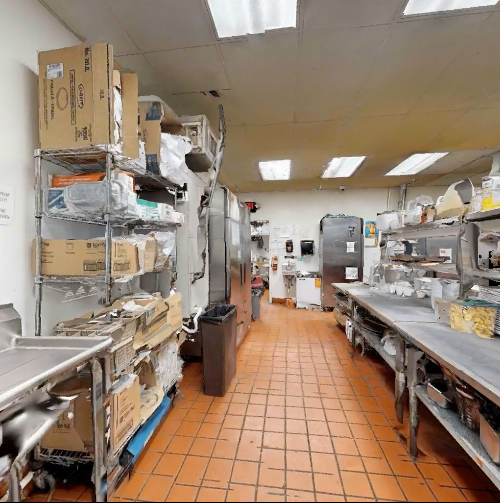}
    \includegraphics[width=0.20\textwidth,valign=m]{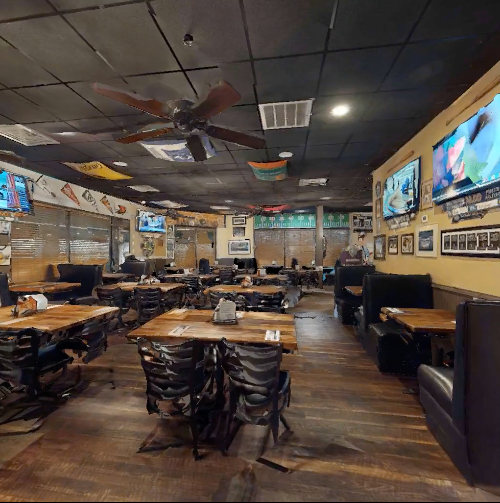} \\
    \includegraphics[trim={0 3cm 0 3cm},clip,width=0.30\textwidth,valign=m]{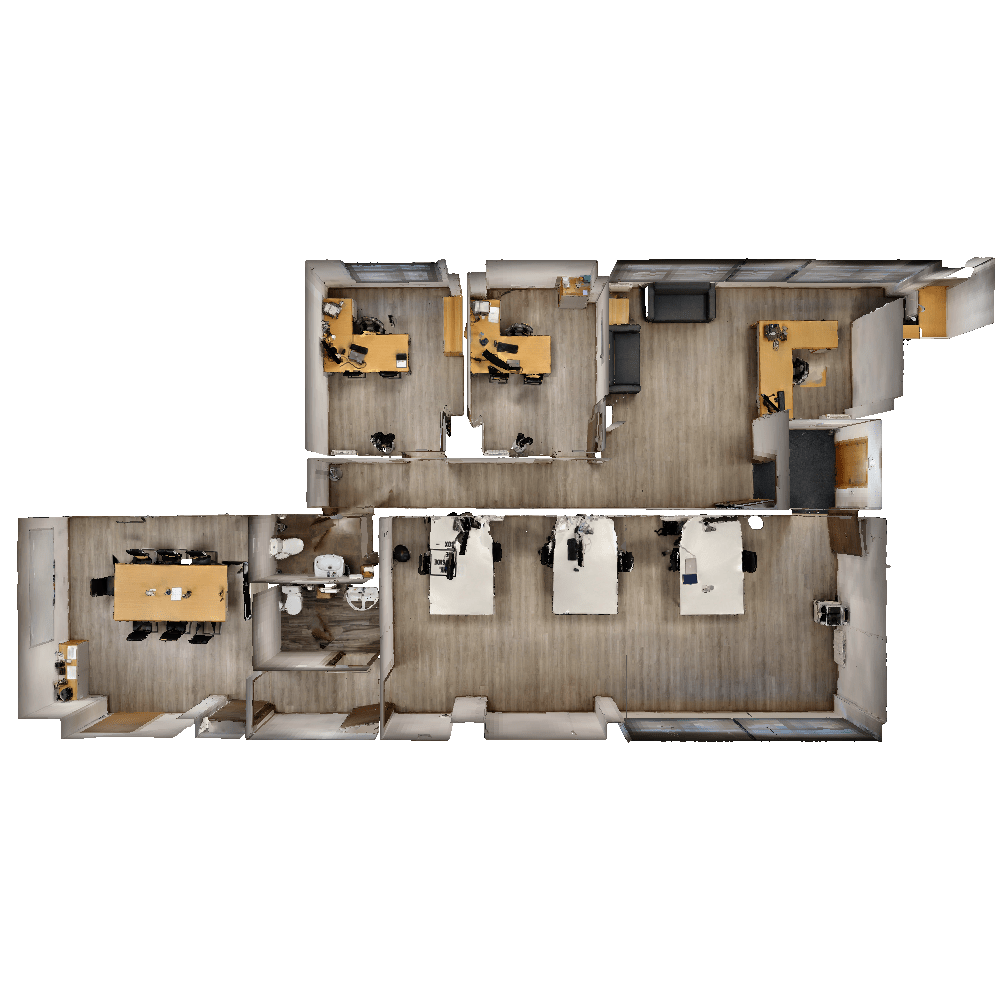}
    \includegraphics[trim={0 3cm 0 3cm},clip,width=0.30\textwidth,valign=m]{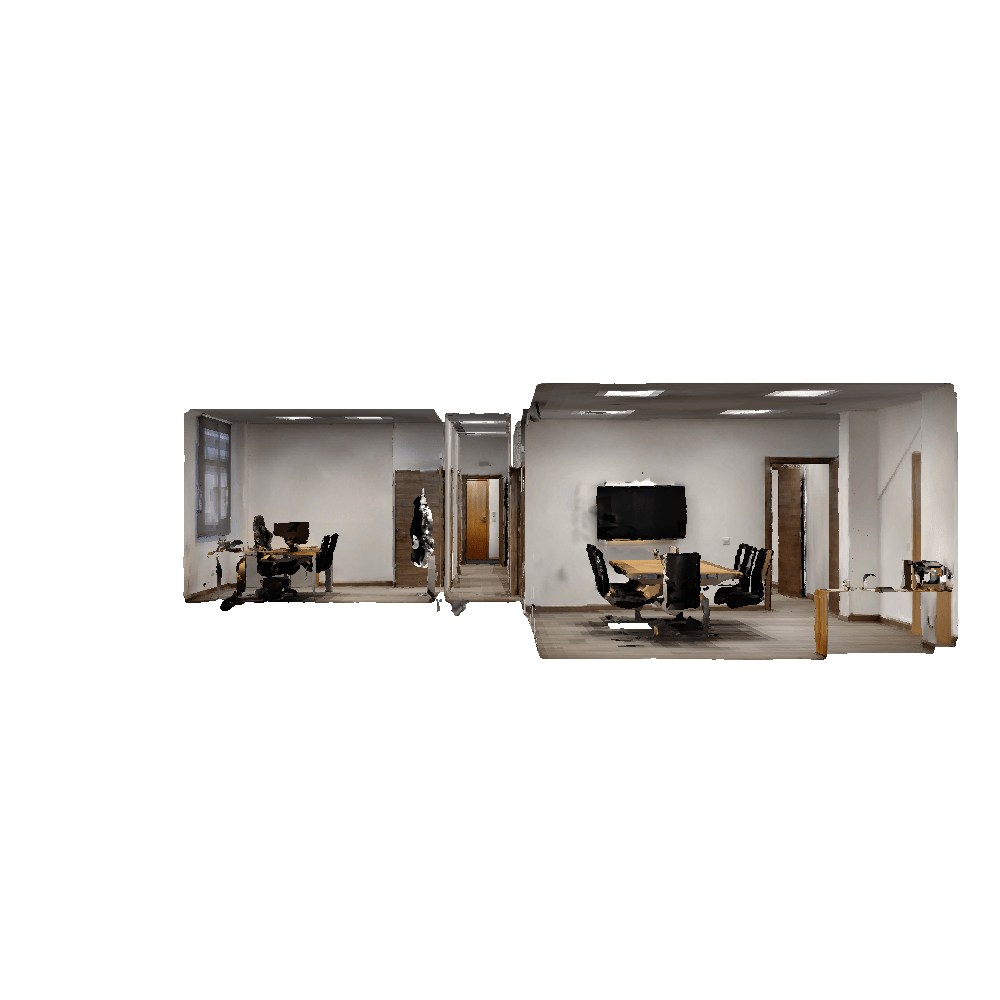}
    \includegraphics[width=0.20\textwidth,valign=m]{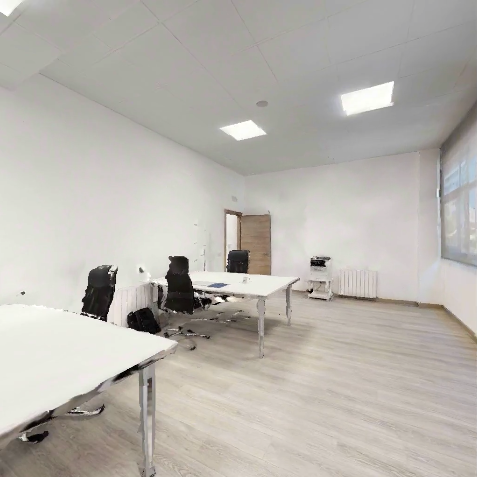}
    \includegraphics[width=0.20\textwidth,valign=m]{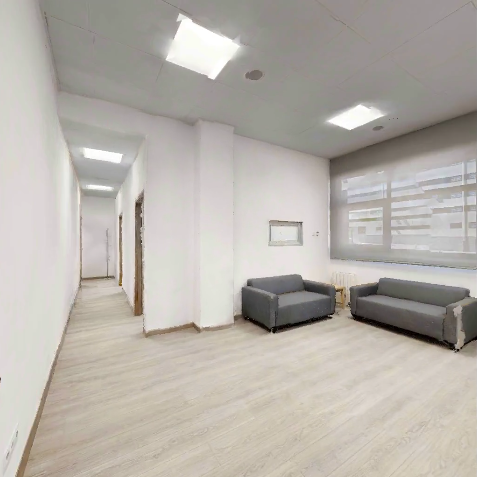} \\
    \includegraphics[trim={0 3cm 0 3cm},clip,width=0.30\textwidth,valign=m]{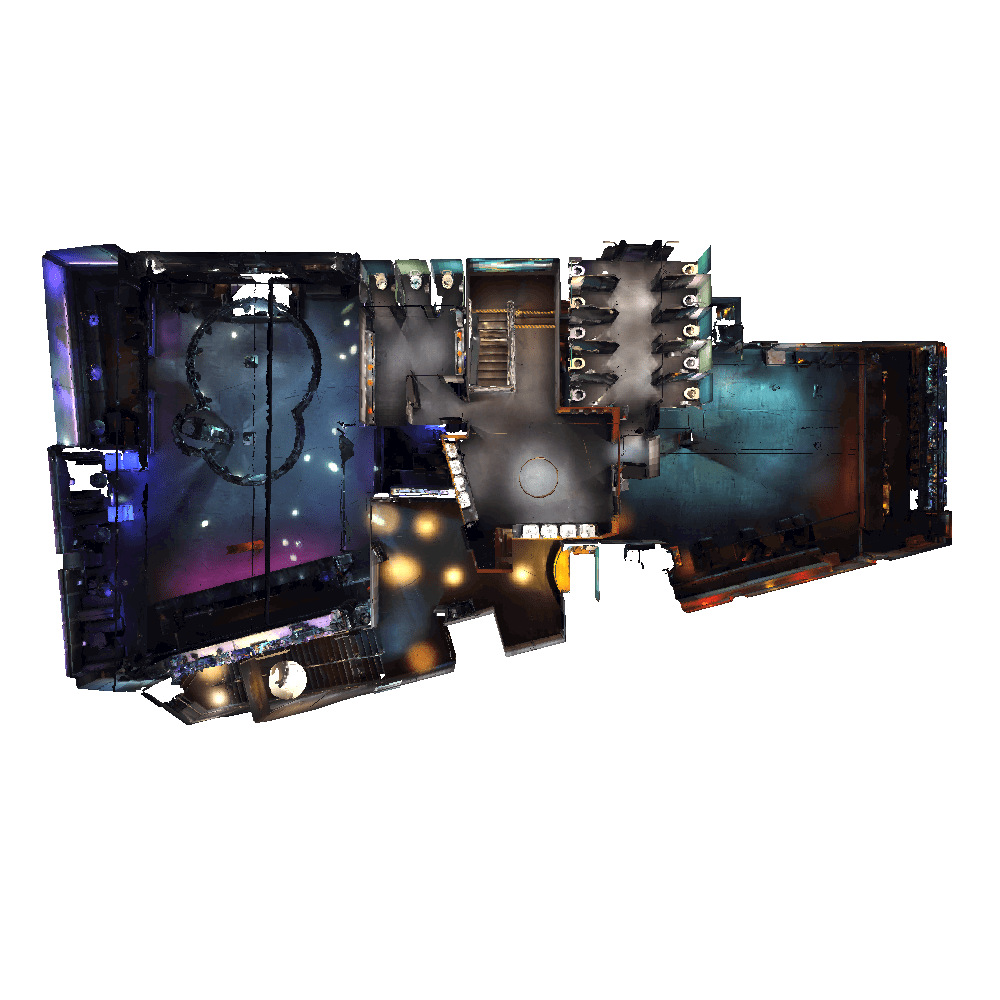}
    \includegraphics[trim={0 3cm 0 3cm},clip,width=0.30\textwidth,valign=m]{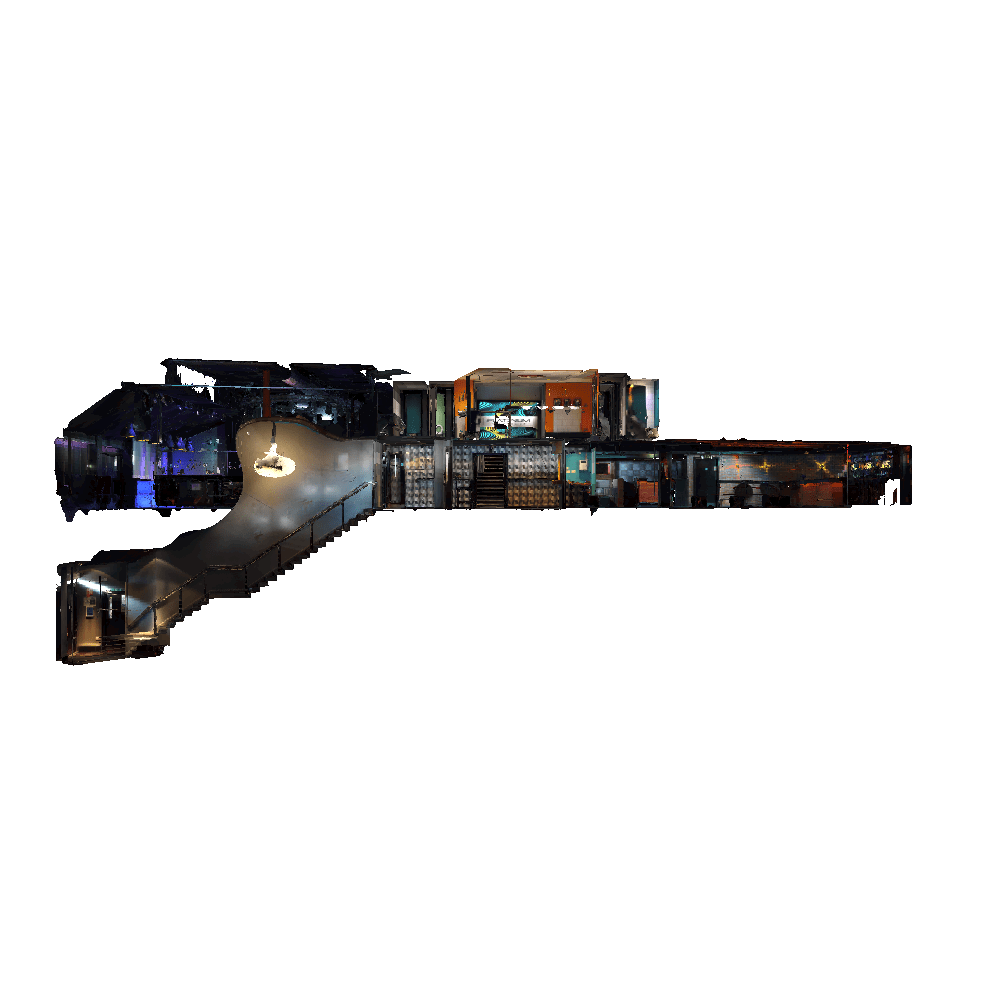}
    \includegraphics[width=0.20\textwidth,valign=m]{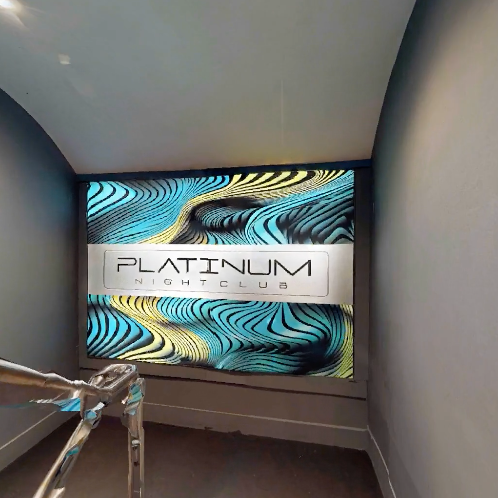}
    \includegraphics[width=0.20\textwidth,valign=m]{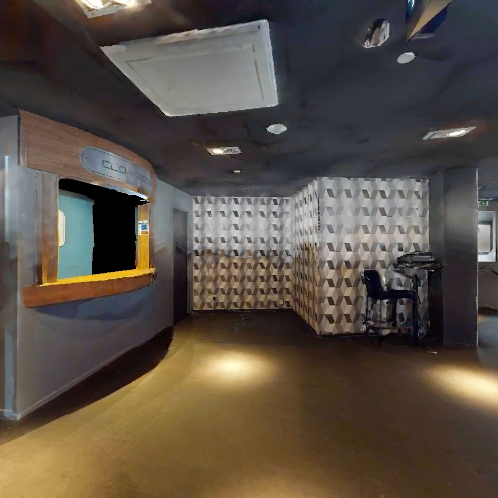} \\
    \includegraphics[trim={0 3cm 0 3cm},clip,width=0.30\textwidth,valign=m]{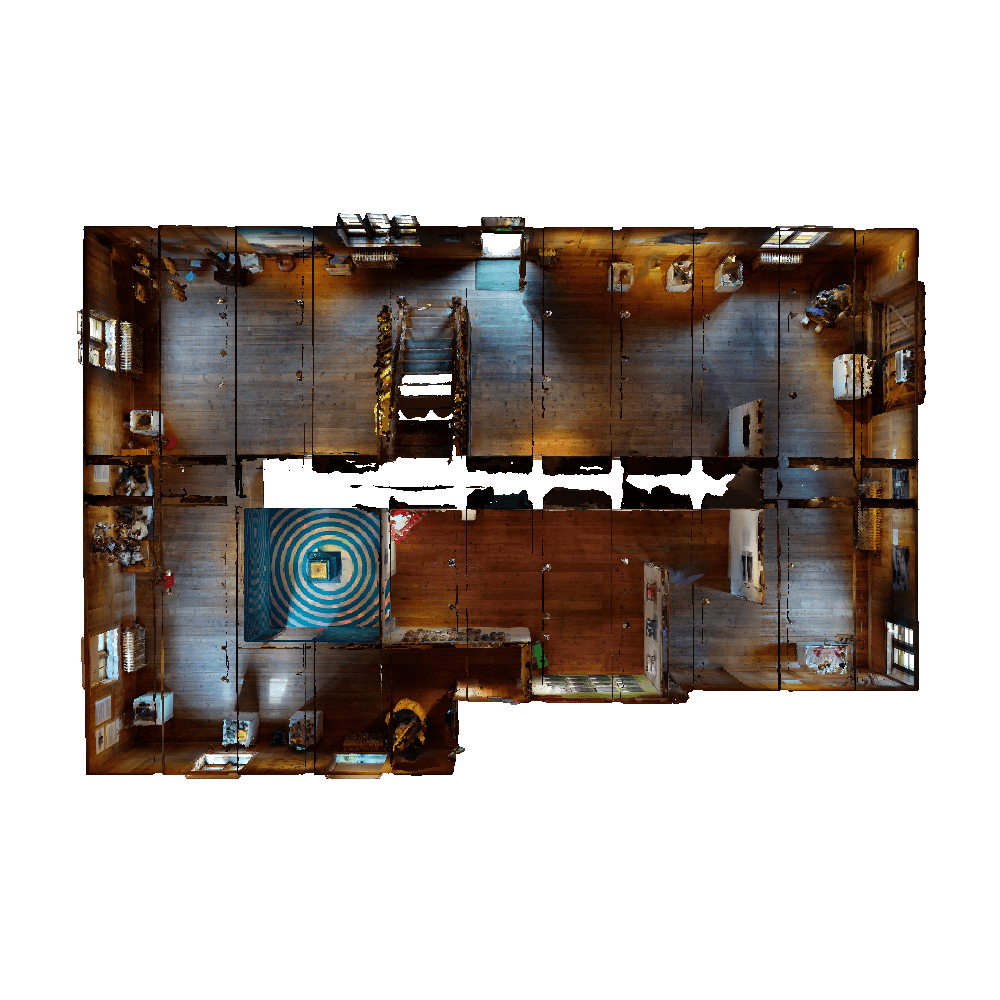}
    \includegraphics[trim={0 3cm 0 3cm},clip,width=0.30\textwidth,valign=m]{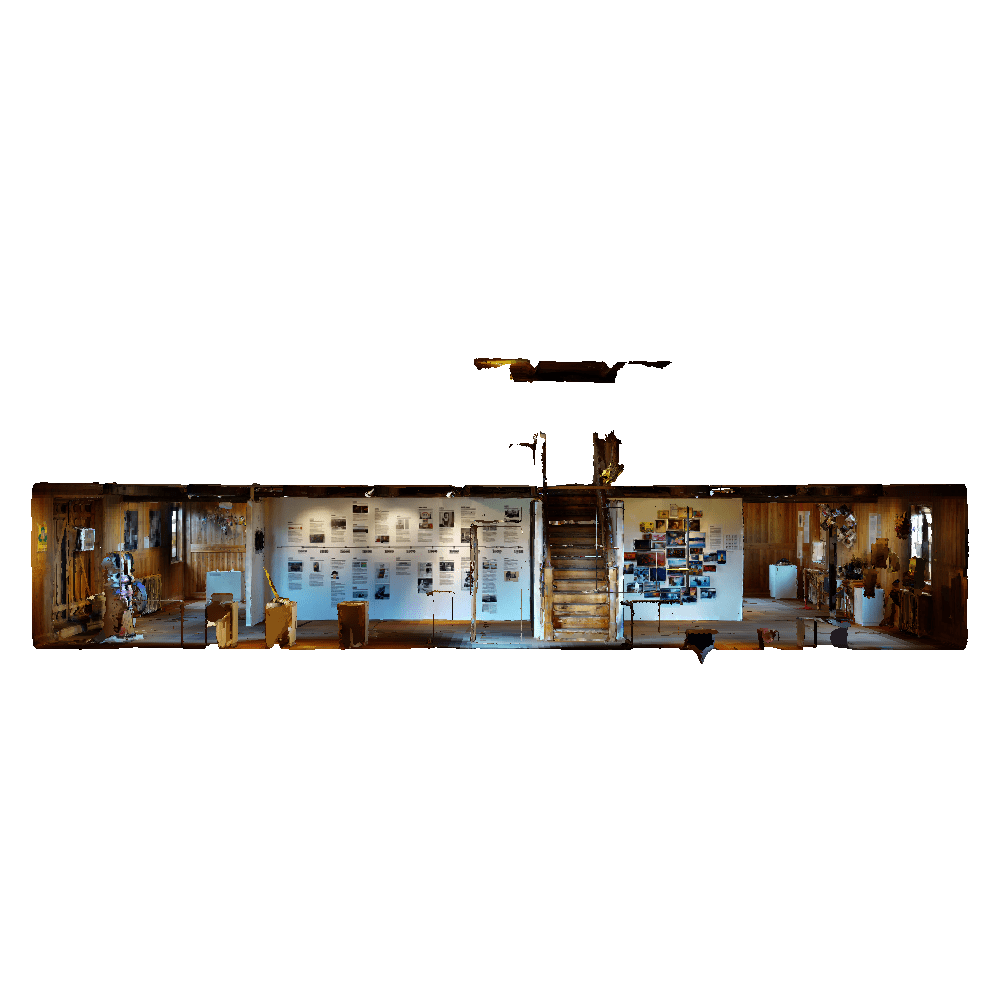}
    \includegraphics[width=0.20\textwidth,valign=m]{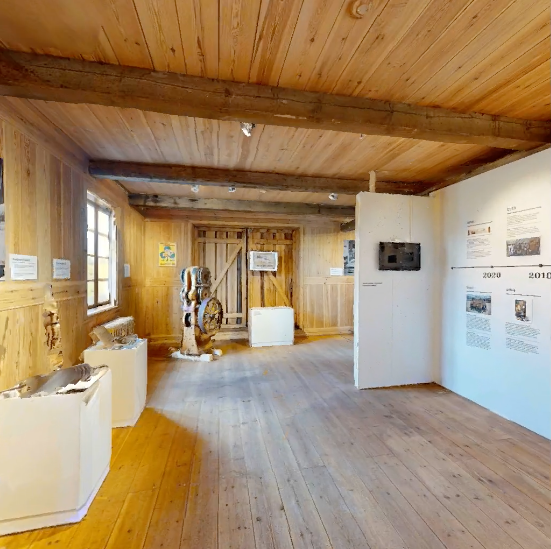}
    \includegraphics[width=0.20\textwidth,valign=m]{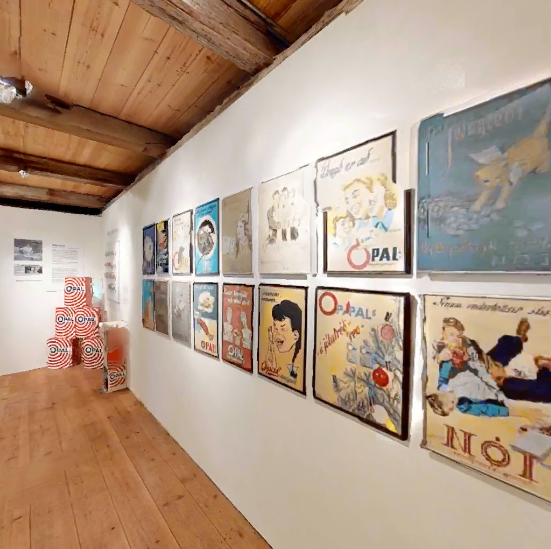} \\
    \includegraphics[trim={3cm 0 3cm 0},clip,height=0.30\textwidth,valign=m,angle=90,origin=c]{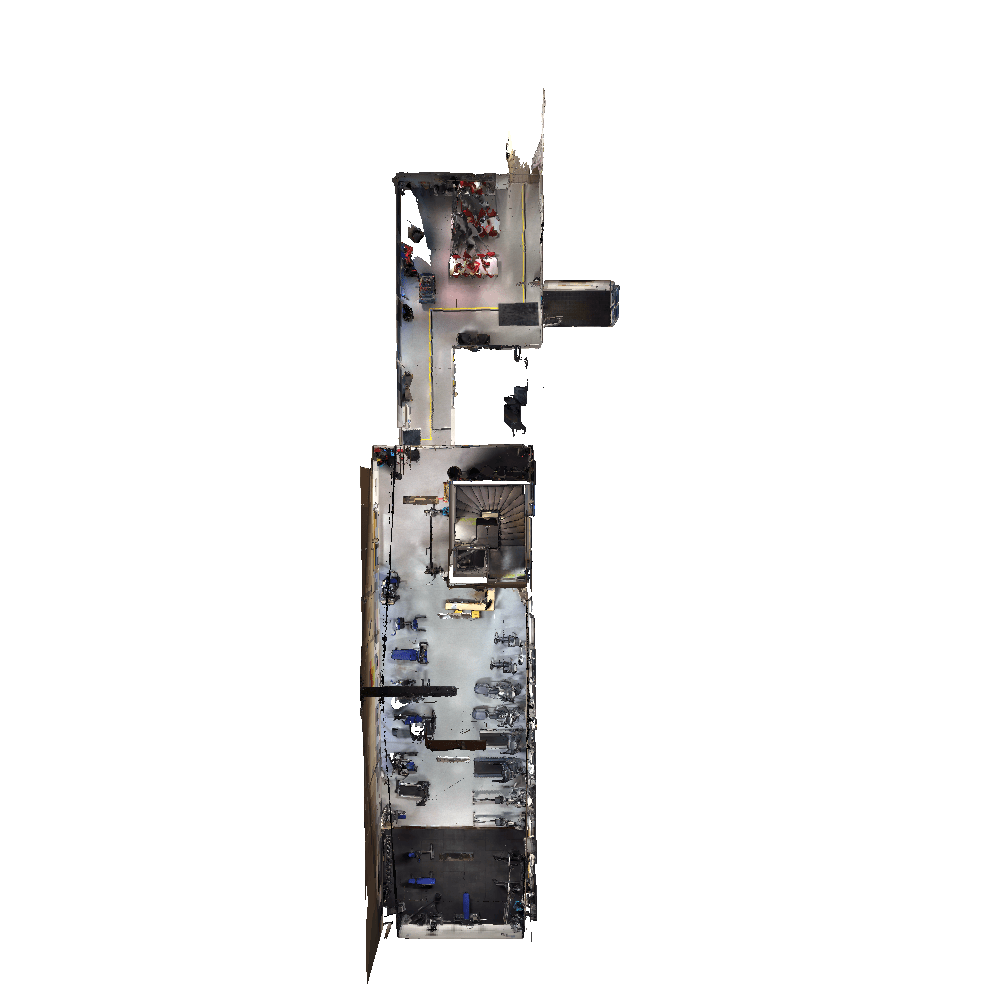}
    \includegraphics[trim={0 3cm 0 3cm},clip,width=0.30\textwidth,valign=m]{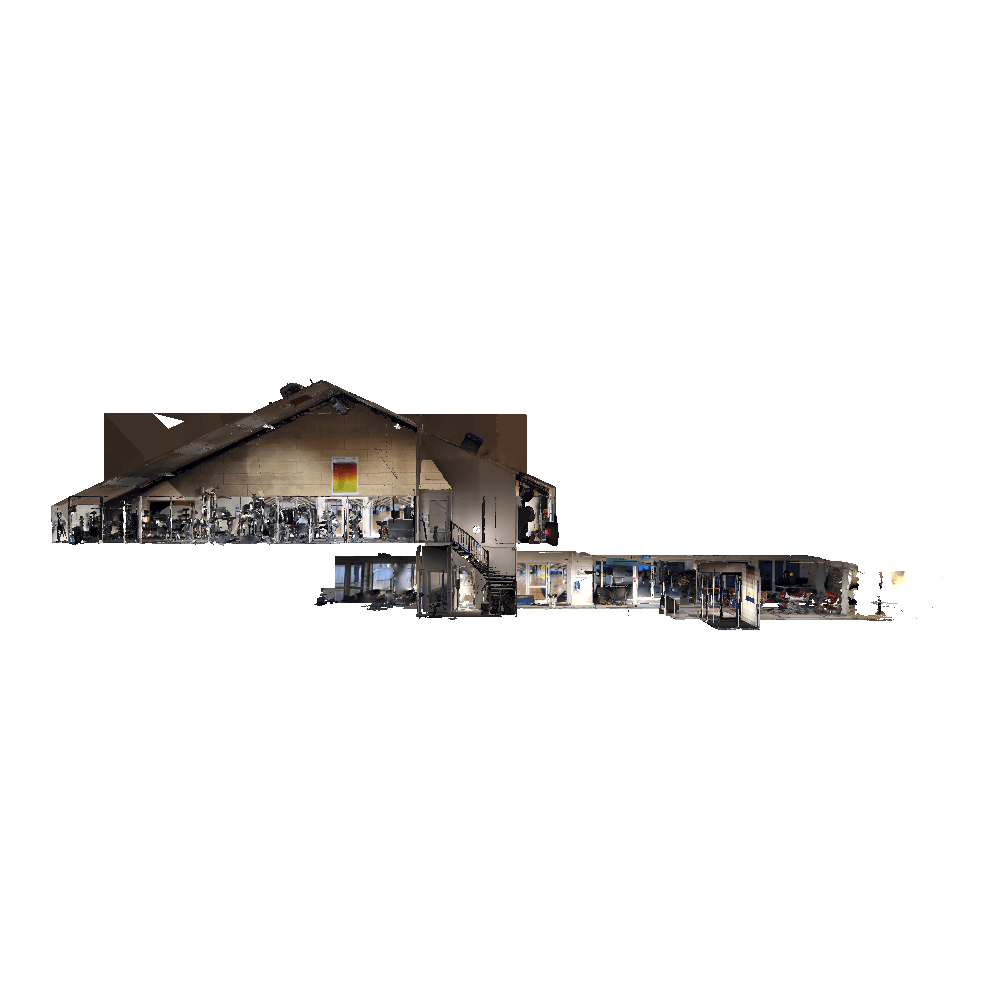}
    \includegraphics[width=0.20\textwidth,valign=m]{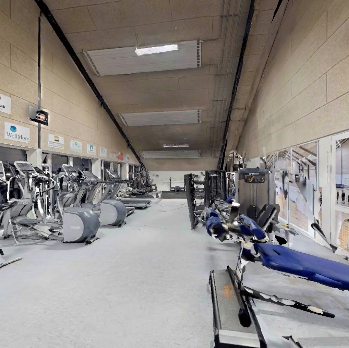}
    \includegraphics[width=0.20\textwidth,valign=m]{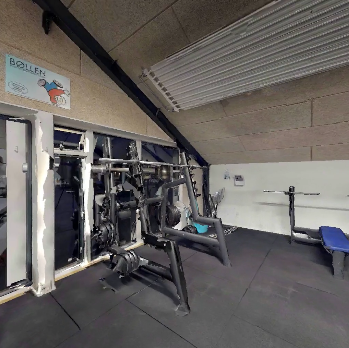} \\
    \end{tabular}
}
\captionof{figure}{Five examples of diverse scenes from the HM3D dataset. From left to right in each row: top-down view, cross section view, and two egocentric views from navigable positions in the scene. The dataset contains diverse scans such as restaurants (row 1), office buildings (row 2), a night club (row 3), art studio (row 4), and gyms (row 5).}
\vspace*{-0.3in}
\label{suppfig:hm3d_examples_2}
\end{table}
\FloatBarrier
\section{PointNav qualitative results}
\label{suppsec:pointnav_qualitative}

We show sample PointNav episodes of the HM3D agents in \Cref{suppfig:pointnav_qual_depth} and \Cref{suppfig:pointnav_qual_rgb}. We present the qualitative results in a format similar to~\citet{wijmans2019dd}. The episodes are categorized based on the difficulty (i.e., the geodesic distance b/w start and goal), and the agent performance (in SPL).

\begin{table}[t]
\resizebox{\textwidth}{!}{
    \centering
    \begin{tabularx}{\linewidth}{Y|YYYY}
    \toprule
    Geodesic distance (rows) / SPL (columns) & $0.00$ - $0.50$ & $0.50$ - $0.90$ & $0.90$ - $0.95$ & $0.95$ - $1.00$ \\\midrule
    $0.0$ - $5.0$ &
    \includegraphics[width=0.20\textwidth,valign=m]{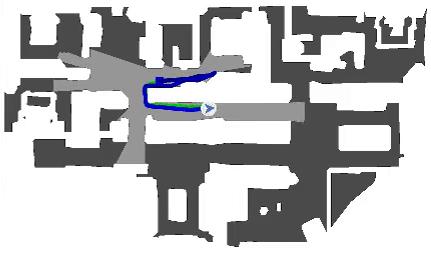} &
    \includegraphics[width=0.20\textwidth,valign=m]{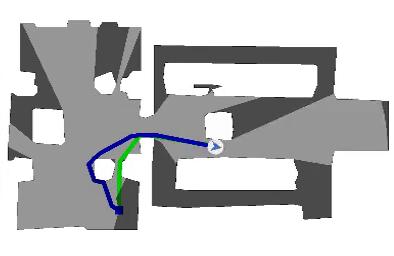} &
    \includegraphics[width=0.20\textwidth,valign=m]{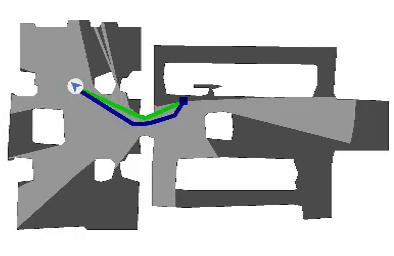} &
    \includegraphics[width=0.20\textwidth,valign=m]{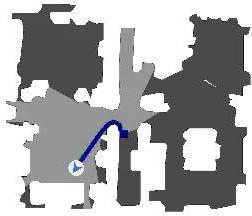} \\
    $5.0$ - $10.0$ &
    \includegraphics[width=0.20\textwidth,valign=m]{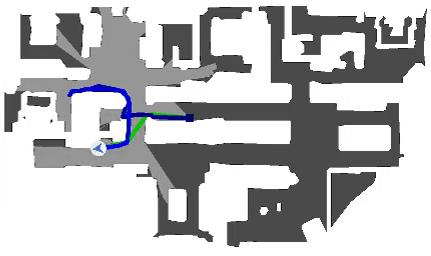} &
    \includegraphics[width=0.20\textwidth,valign=m]{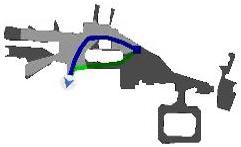} &
    \includegraphics[width=0.20\textwidth,valign=m]{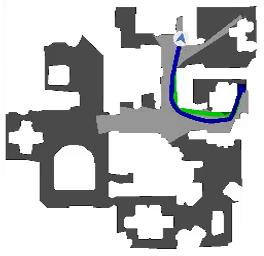} &
    \includegraphics[width=0.20\textwidth,valign=m]{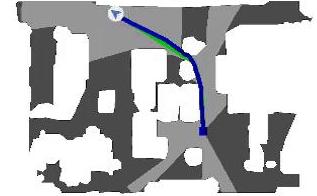} \\
    $10.0$ - $15.0$ &
    \includegraphics[width=0.20\textwidth,valign=m]{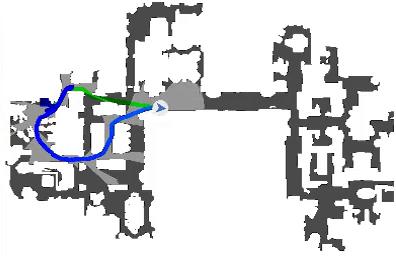} &
    \includegraphics[width=0.20\textwidth,valign=m]{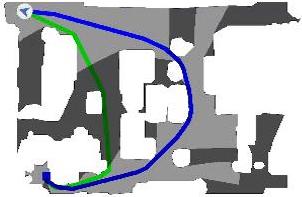} &
    \includegraphics[width=0.20\textwidth,valign=m]{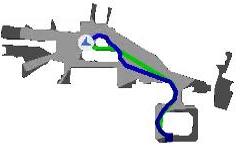} &
    \includegraphics[width=0.20\textwidth,valign=m]{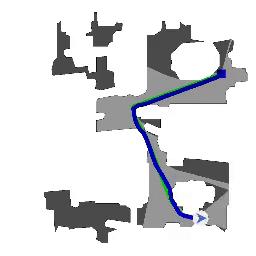} \\
    $15.0$ - $20.0$ &
    \includegraphics[width=0.20\textwidth,valign=m]{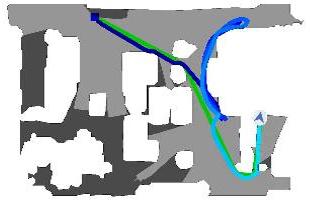} &
    \includegraphics[width=0.20\textwidth,valign=m]{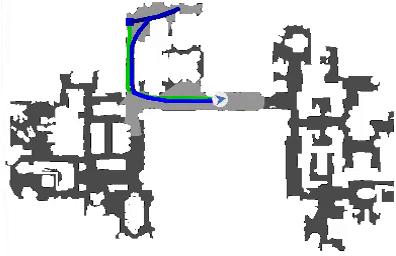} &
    \includegraphics[width=0.20\textwidth,valign=m]{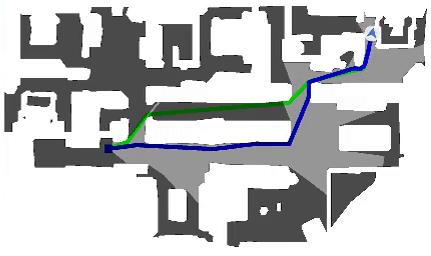} &
    \includegraphics[width=0.20\textwidth,valign=m]{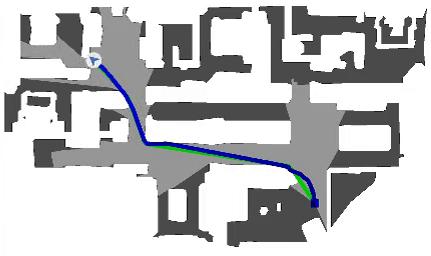} \\
    $20.0$ - $25.0$ &
    \includegraphics[width=0.20\textwidth,valign=m]{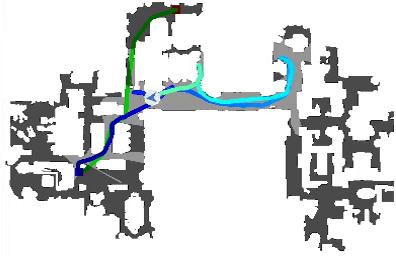} &
    \includegraphics[width=0.20\textwidth,valign=m]{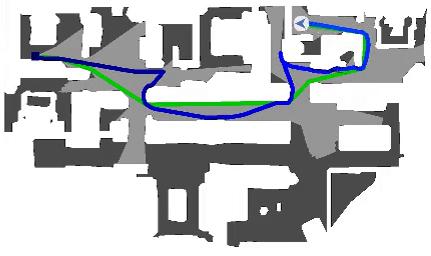} &
    \includegraphics[width=0.20\textwidth,valign=m]{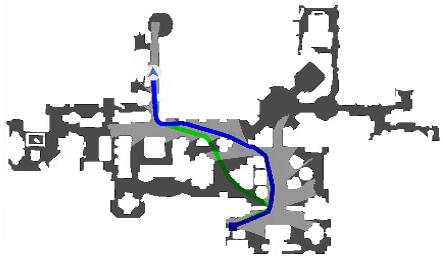} &
    \includegraphics[width=0.20\textwidth,valign=m]{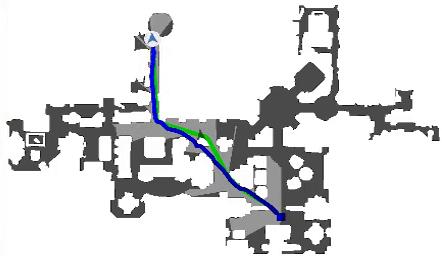} \\
    $25.0$ - $30.0$ &
    \includegraphics[width=0.20\textwidth,valign=m]{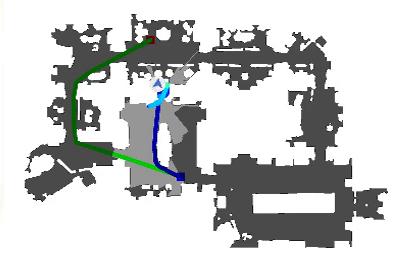} &
    \includegraphics[width=0.20\textwidth,valign=m]{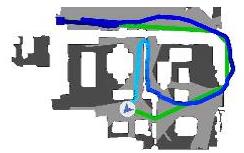} &
    \includegraphics[width=0.20\textwidth,valign=m]{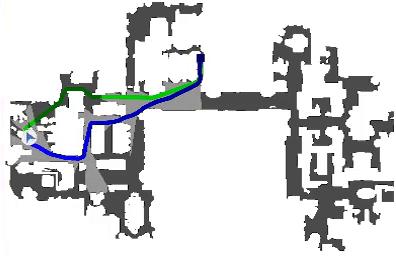} &
    \includegraphics[width=0.20\textwidth,valign=m]{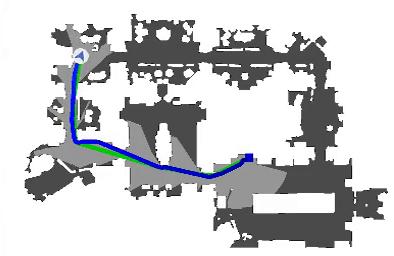} \\\bottomrule
    \end{tabularx}
}
\vspace{0.05in}
\captionof{figure}{\textbf{PointNav with depth sensor:} Example episodes for the HM3D-agent with depth inputs broken down by geodesic distance between agent’s start and
goal locations (on rows) vs SPL achieved by the agent (on columns). Gray represents navigable regions on the
map while white is non-navigable. The agent starts at the blue square and navigates to the red
square. The green line shows the shortest path on the map (or oracle navigation). The blue line
shows the agent’s trajectory. The SPL score is higher if the blue trajectory closely matches the green trajectory.}
\vspace*{-0.3in}
\label{suppfig:pointnav_qual_depth}
\end{table}

\begin{table}[t]
\resizebox{\textwidth}{!}{
    \centering
    \begin{tabularx}{\linewidth}{Y|YYYY}
    \toprule
    Geodesic distance (rows) / SPL (columns) & $0.00$ - $0.50$ & $0.50$ - $0.90$ & $0.90$ - $0.95$ & $0.95$ - $1.00$ \\\midrule
    $0.0$ - $5.0$ &
    \includegraphics[width=0.20\textwidth,valign=m]{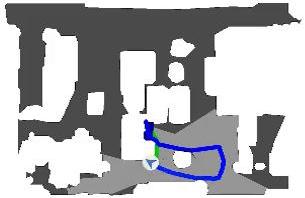} &
    \includegraphics[width=0.20\textwidth,valign=m]{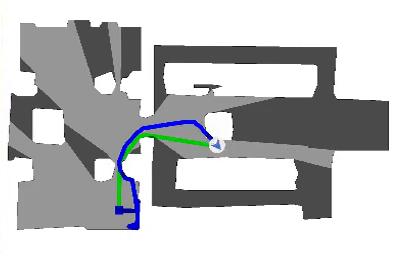} &
    \includegraphics[width=0.20\textwidth,valign=m]{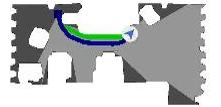} &
    \includegraphics[width=0.20\textwidth,valign=m]{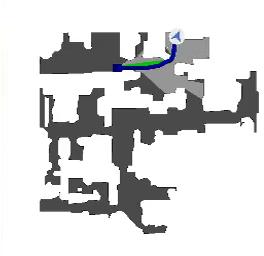} \\
    $5.0$ - $10.0$ &
    \includegraphics[width=0.20\textwidth,valign=m]{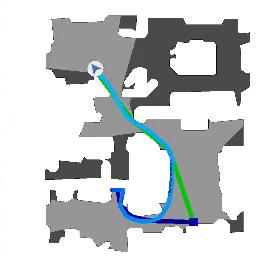} &
    \includegraphics[width=0.20\textwidth,valign=m]{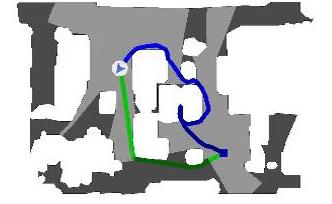} &
    \includegraphics[width=0.20\textwidth,valign=m]{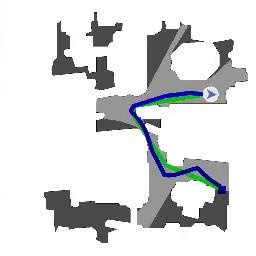} &
    \includegraphics[width=0.20\textwidth,valign=m]{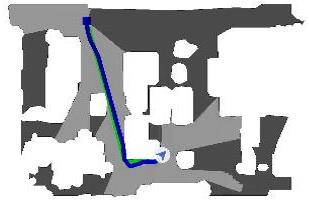} \\
    $10.0$ - $15.0$ &
    \includegraphics[width=0.20\textwidth,valign=m]{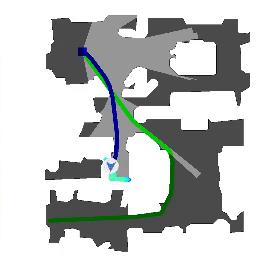} &
    \includegraphics[width=0.20\textwidth,valign=m]{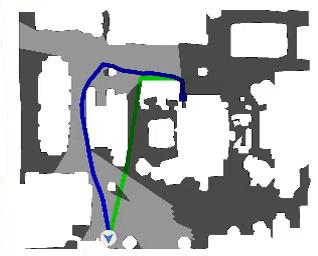} &
    \includegraphics[width=0.20\textwidth,valign=m]{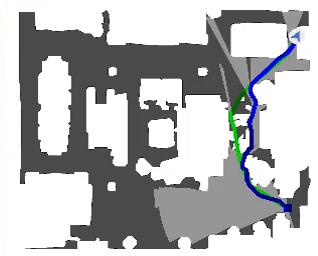} &
    \includegraphics[width=0.20\textwidth,valign=m]{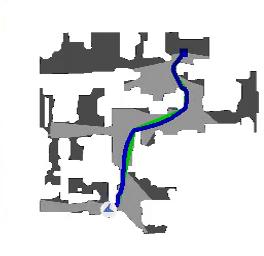} \\
    $15.0$ - $20.0$ &
    \includegraphics[width=0.20\textwidth,valign=m]{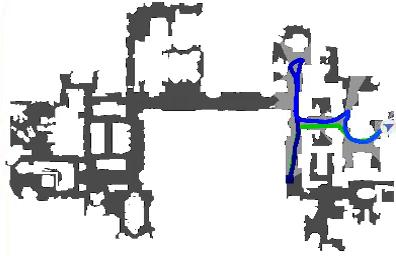} &
    \includegraphics[width=0.20\textwidth,valign=m]{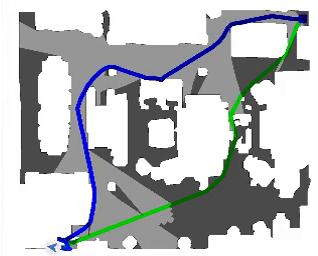} &
    \includegraphics[width=0.20\textwidth,valign=m]{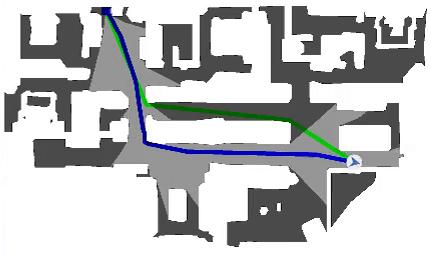} &
    \includegraphics[width=0.20\textwidth,valign=m]{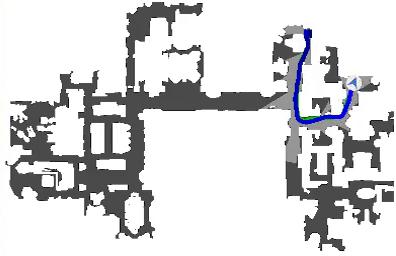} \\
    $20.0$ - $25.0$ &
    \includegraphics[width=0.20\textwidth,valign=m]{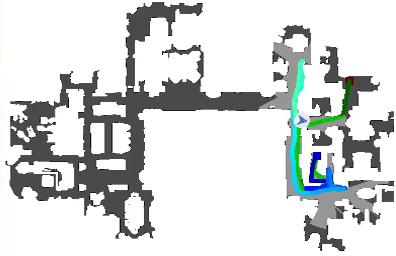} &
    \includegraphics[width=0.20\textwidth,valign=m]{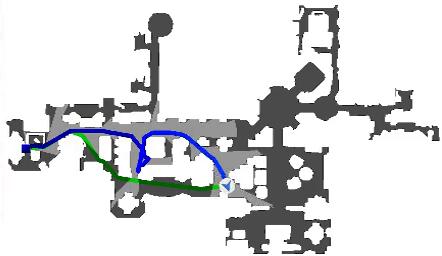} &
    \includegraphics[width=0.20\textwidth,valign=m]{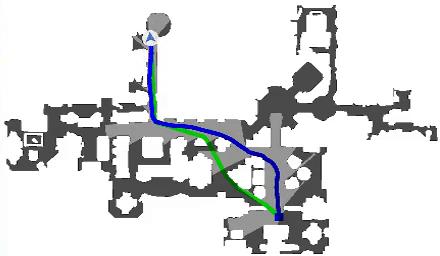} &
    \includegraphics[width=0.20\textwidth,valign=m]{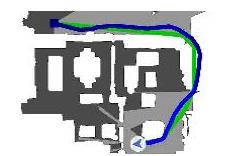} \\
    $25.0$ - $30.0$ &
    \includegraphics[width=0.20\textwidth,valign=m]{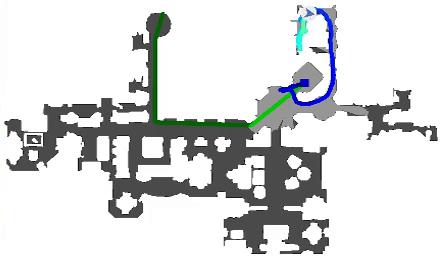} &
    \includegraphics[width=0.20\textwidth,valign=m]{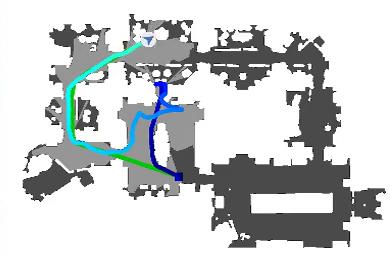} &
    \includegraphics[width=0.20\textwidth,valign=m]{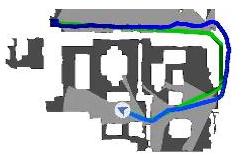} &
    \includegraphics[width=0.20\textwidth,valign=m]{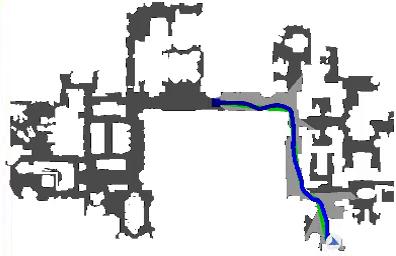} \\\bottomrule
    \end{tabularx}
}
\vspace{0.05in}
\captionof{figure}{\textbf{PointNav with RGB sensor:} Example episodes for the HM3D-agent with RGB inputs broken down by geodesic distance between agent’s start and
goal locations (on rows) vs SPL achieved by the agent (on columns). Gray represents navigable regions on the
map while white is non-navigable. The agent starts at the blue square and navigates to the red
square. The green line shows the shortest path on the map (or oracle navigation). The blue line
shows the agent’s trajectory.  The SPL score is higher if the blue trajectory closely matches the green trajectory.}
\vspace*{-0.3in}
\label{suppfig:pointnav_qual_rgb}
\end{table}

\end{document}